\documentclass{article} 
\usepackage{iclr2025_conference,times}

\usepackage{hyperref}
\usepackage{url}

\usepackage{pifont}
\usepackage{booktabs}
\usepackage{multirow}
\usepackage{graphicx}
\usepackage{amsmath}
\usepackage{subcaption}
\usepackage{blindtext}
\usepackage{float}
\usepackage{amssymb}
\usepackage{pifont}
\newcommand{\cmark}{\ding{51}}%
\newcommand{\xmark}{\ding{55}}%

\usepackage[utf8]{inputenc} 
\usepackage[T1]{fontenc}    
\usepackage{url}            
\usepackage{booktabs}       
\usepackage{amsfonts}       
\usepackage{nicefrac}       
\usepackage{microtype}      
\usepackage{xcolor}         
\usepackage[font={small}]{caption}

\title{Learning to Generate Diverse Pedestrian Movements from Web Videos with Noisy Labels}

\author{Zhizheng Liu, Joe Lin, Wayne Wu, Bolei Zhou
\\Department of Computer Science,
        University of California, Los Angeles
}

\iclrfinalcopy

\begin{document}

\maketitle
\begin{figure}[h]
    \centering
    \includegraphics[width=0.9\linewidth]{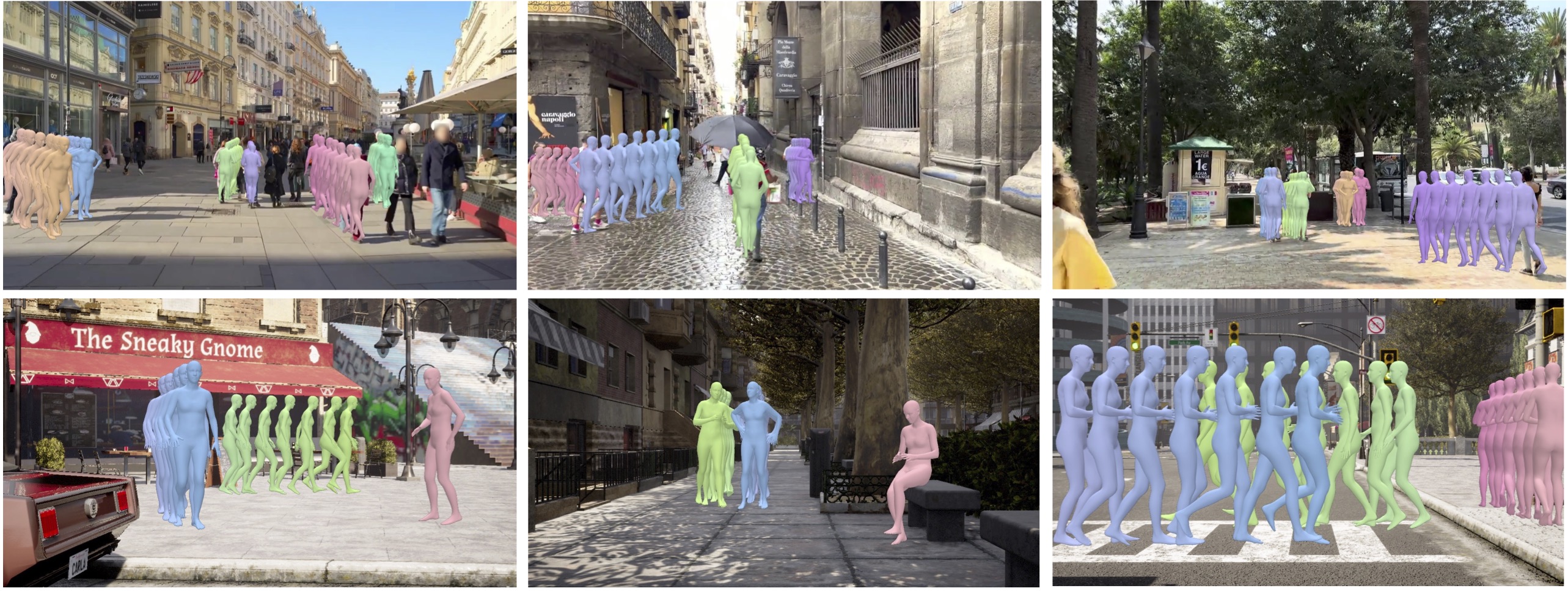}
    \caption{\textbf{Pedestrian Movement Generation.} Our method can generate diverse pedestrian movements in real-world (top row) and simulated (bottom row) urban environments.
    }
 
    \label{fig:teaser}
    
\end{figure}

\begin{abstract}
Understanding and modeling pedestrian movements in the real world is crucial for applications like motion forecasting and scene simulation. Many factors influence pedestrian movements, such as scene context, individual characteristics, and goals, which are often ignored by the existing human generation methods.
Web videos contain natural pedestrian behavior and rich motion context, but annotating them with pre-trained predictors leads to noisy labels. 
In this work, we propose learning diverse pedestrian movements from web videos. 
We first curate a large-scale dataset called CityWalkers that captures diverse real-world pedestrian movements in urban scenes.
Then, based on CityWalkers,  we propose a generative model called PedGen for diverse pedestrian movement generation. PedGen introduces automatic label filtering to remove the low-quality labels and a mask embedding to train with partial labels. It also contains a novel context encoder that lifts the 2D scene context to 3D and can incorporate various context factors in generating realistic pedestrian movements in urban scenes. 
Experiments show that PedGen outperforms existing baseline methods for pedestrian movement generation by learning from noisy labels and incorporating the context factors. In addition, PedGen achieves zero-shot generalization in both real-world and simulated environments. The code, model, and data will be made publicly available at \url{https://genforce.github.io/PedGen/}.
\end{abstract}
\section{Introduction}
In a bustling city, office workers rush through the crosswalk to nearby buildings, tourists wander around along storefronts looking at items on display, and friends sit and enjoy coffee on the outdoor patio. Pedestrians are essential participants in urban spaces. Their movements represent the social lives and interactions with the surrounding environments. Understanding and modeling pedestrian movements is critical to many applications. For example, city designers simulate pedestrian movements to optimize public areas and transportation systems~\citep{mehta2008walkable}; forecasting the future pedestrian path is crucial for the safe deployment of autonomous vehicles~\citep{lyssenko2021evaluation}.

Generating diverse and realistic pedestrian movements remains challenging. Multiple context factors affect pedestrian movements in the scenes. The first factor is the surrounding environment. As people constantly interact with the environment, it is important to model the scene context where the interaction happens. For example, objects like trash bins, plants, and other pedestrians in the scene can influence the walking behavior of a pedestrian~\citep{daamen2003experimental}, while the space types like crosswalks and waiting zones decide the overall movement pattern~\citep{sime1995crowd}.  The second factor is the individual characteristics. Studies have shown that walking speed and gait are influenced by age~\citep{ostrosky1994comparison}, gender~\citep{yamasaki1991sex}, and body weight~\citep{heglund1988speed} and can also reflect the pedestrian's fitness level~\citep{dridi2015list}. The last factor is the goal of the pedestrian, which decides the walking route in the scene~\citep{hoogendoorn2004pedestrian}. 

Existing human motion generation research mainly focuses on the breadth of activities~\citep{mahmood2019amass}, but few have studied generating natural real-world pedestrian movements in diverse scene contexts. 
For example, existing outdoor human motion datasets have limited scenes and human subjects with unnatural motion performed by actors with specific instructions~\citep{kaufmann2023emdb, Dai_2023_CVPR}. Moreover, current motion generation methods~\citep{guo2020action2motion, tevet2022human} lean toward generating complex motions from clean MoCap data, where many motion categories are rare in daily scenes. Few have considered learning the diverse motion contexts from noisy labels.

Web videos, captured by many people walking and touring in different cities worldwide and shared on the YouTube website, contain diverse scene contexts and pedestrian movements in the most natural forms. 
However, labeling pedestrian movements in these web videos with pretrained predictors leads to inevitable label noise.
How to harness the web data with large noise yet rich context becomes the pivot of modeling and generating diverse pedestrian movements.
To this end, we first collect \textit{CityWalkers}, a \textit{large-scale} real-world dataset containing pedestrians in urban scenes, annotated with pseudo-labels by off-the-shelf 4D human motion estimation models. 
CityWalkers captures diverse real-world pedestrian movements regarding various moving speeds, gaits, headings, and local motions. Each movement is also paired with labels of context factors, such as the pedestrian's body shape, route destinations, and the environment's semantics and geometry.

We then develop a new diffusion-based generative model \textit{PedGen} for learning context-aware pedestrian movements with the noisy pseudo-labels from the CityWalkers dataset. PedGen has two key designs: 
1) To mitigate the anomaly and incomplete labels from pseudo-labeling techniques, PedGen adopts a data iteration strategy to identify and remove low-quality labels from the dataset automatically and a motion mask embedding to train with partial labels;
2) To model the important context factors, PedGen considers the surrounding environment, the individual characteristics, and the goal points as input conditions to generate realistic and long-term pedestrian movements in urban scenes.
As web videos only contain scene context labels in 2D, we propose a novel Context Encoder that can lift the environment context from 2D images into a 3D local scene representation with geometry and semantic information and also encode the other context factors to help generate realistic and long-term 3D pedestrian movements.
We show some randomly sampled results of PedGen in Fig.~\ref{fig:teaser}.

Experiment results on the CityWalkers validation set, the real-world Waymo open dataset~\citep{sun2020scalability} and CARLA simulator~\citep{dosovitskiy2017carla} show that PedGen can predict more realistic and accurate future pedestrian movements than existing human motion generation methods and achieve better zero-shot generalization by generating high-quality context-aware movements. Additional experiments and ablation studies demonstrate the effectiveness of PedGen in addressing noisy labels and incorporating the key context factors. It enables the application of forecasting pedestrian movements in the real world and populating simulated environments with realistic pedestrians. We summarize \textit{our contributions} as follows:
1) A new task of context-aware pedestrian movement generation from web videos with unique challenges in dealing with label noises and modeling various motion contexts.
2) A new large-scale real-world pedestrian movement dataset CityWalkers with pseduo-labels of diverse pedestrian movements and motion contexts.
3) The context-aware generative model PedGen that can learn from noisy pseudo-labels to generate diverse pedestrian movements.

\section{Related Work}
\paragraph{Pedestrian Movement Analysis.}
Pedestrian behaviors have been extensively studied in transportation and social science. A hierarchical structure is defined for pedestrian behavior analysis~\citep{hoogendoorn2004pedestrian, feng2021data} from the high-level strategic behavior, the middle-level tactical behavior, to the low-level operational behavior. Pedestrian movements belong to operational behaviors, where pedestrians continuously make short-term movement decisions on their route to respond to their immediate environment~\citep{daamen2002modelling, duives2016analysis}. Many works analyze different factors influencing pedestrian movements~\citep{dridi2015list}. Some critical environmental factors include types of space~\citep{sime1995crowd}, objects in the environment~\citep{daamen2003experimental}, and movement of other pedestrians~\citep{van2016influence}. Pedestrian movements also depend on their biometric data, like age~\citep{ostrosky1994comparison}, gender~\citep{yamasaki1991sex}, and body size~\citep{heglund1988speed}.
Different from most existing works in pedestrian movement analysis that collect data from field observations~\citep{shields2000study} or controlled experiments~\citep{haghani2018crowd} and analyze them using statistical approaches~\citep{tong2023investigation}, we extract pedestrian movements from web videos and learn a generative model to facilitate pedestrian movement modeling.

\paragraph{Human Motion Datasets.}
AMASS~\citep{mahmood2019amass} is one of the most popular human motion datasets with diverse motions annotated with SMPL~\citep{loper2023smpl} parameters. The main issue of AMASS is the lack of context information related to the motion. Later datasets have focused on adding more human subjects~\citep{cheng2023dna}, text descriptions~\citep{Guo_2022_CVPR}, human-object interactions~\citep{bhatnagar2022behave} and human-scene interactions~\citep{hassan2019resolving,hassan2021stochastic, huang2022capturing, jiang2024scaling} to the labels. Nevertheless, most of these datasets are captured in a controlled environment with the subjects asked to follow specific action instructions, and many motions, such as jump jacks and martial arts, are rarely seen in urban scenes. In-the-wild videos serve as a more suitable source for studying human movement with their richness in individuals and environments.
Another line of work~\citep{von2018recovering,  guzov2021human, Dai_2023_CVPR, kaufmann2023emdb} focuses on collecting human motion in outdoor places from in-the-wild videos, but their additional sensor requirements like IMUs limit their scalability to collect large-scale datasets.
\cite{zhu2021gait, zheng2022gait} propose benchmarks for gait recognition in the wild, but these datasets lack diverse scene contexts.
Some other datasets~\citep{vendrow2023jrdb, robicquet2016learning} label human social behaviors and trajectories from street videos as key points or bounding boxes. Still, these labels have worse motion granularity than the SMPL parameters for pedestrian movement analysis. Our proposed dataset, CityWalkers, consists of large-scale web videos of pedestrians in diverse urban environments. CityWalkers provides both SMPL movement labels and context pseudo-labels, including the body shape of the pedestrians, their route destinations, and the semantics and geometry of the scene.

\paragraph{Human Motion Generation.}
Human motion generation has been accelerated by large-scale motion datasets and rapid advancements in generative models, with diffusion models~\citep{ho2020denoising} being the most successful architecture for its high generation quality and multi-modal modeling capacity. To generate motion with more fine-grained control, various input conditions have been used, including action labels~\citep{tevet2022human}, texts~\citep{zhang2024motiondiffuse}, audios~\citep{dabral2023mofusion}, and history motions~\citep{chen2023humanmac}. The most relevant motion generation models to ours are the ones that condition on the indoor scenes~\citep{huang2023diffusion, yi2024tesmo, jiang2024scaling}. However, there is a huge gap between indoor and outdoor environments, and generating pedestrian movement requires considering more context factors other than the surrounding environment, like route destinations and pedestrian characteristics. Some works~\citep{tripathi2024humos, xue2024shape} propose to condition the motion on the body shape but do not consider the scene context. ~\citet{rempe2023trace} animate pedestrian movements by generating high-level trajectories and then training a policy to control the pedestrian movements in simulation. Nevertheless, their movement data for training the RL policy comes from a subset of AMASS~\citep{mahmood2019amass} with limited diversity. ~\citet{wang2024pacer+} can generate diverse pedestrian animations while following the given trajectory, but the motion is not learned from real-world pedestrians. ~\citet{shan2023animating} tackle pedestrian movement generation with rule-based approaches by combining path planning algorithms~\citep{treuille2006continuum} and manually designed animations, lacking the diversity and realism of real-world pedestrian movements. On the contrary, our proposed PedGen model learns to generate diverse and realistic pedestrian movements conditioned on the context factors from large-scale real-world data with noisy labels.

\section{Capturing Diverse Real-World Pedestrian Movements }
\label{sec:dataset}
 This section introduces our effort to capture real-world pedestrian movements from web videos. Existing human motion datasets rarely capture natural pedestrian movements, lack diversity in scenes and human subjects, and do not provide the critical context factors of pedestrian movements, such as surrounding environments, individual characteristics, and route destinations. To support the task of context-aware pedestrian movement generation, we construct CityWalkers, a large-scale dataset with real-world pedestrian movements in diverse urban environments annotated by pseudo-labeling techniques. Sec.~\ref{dataset_stats} shows an overview of CityWalkers, and Sec.~\ref{dataset_label} describes our data collection and annotation procedure. Please refer to the Appendix for more details about CityWalkers.

\subsection{CityWalkers Dataset}
\label{dataset_stats}
\begin{figure}[t!]
    \centering
   
    \includegraphics[width=1.0\linewidth]{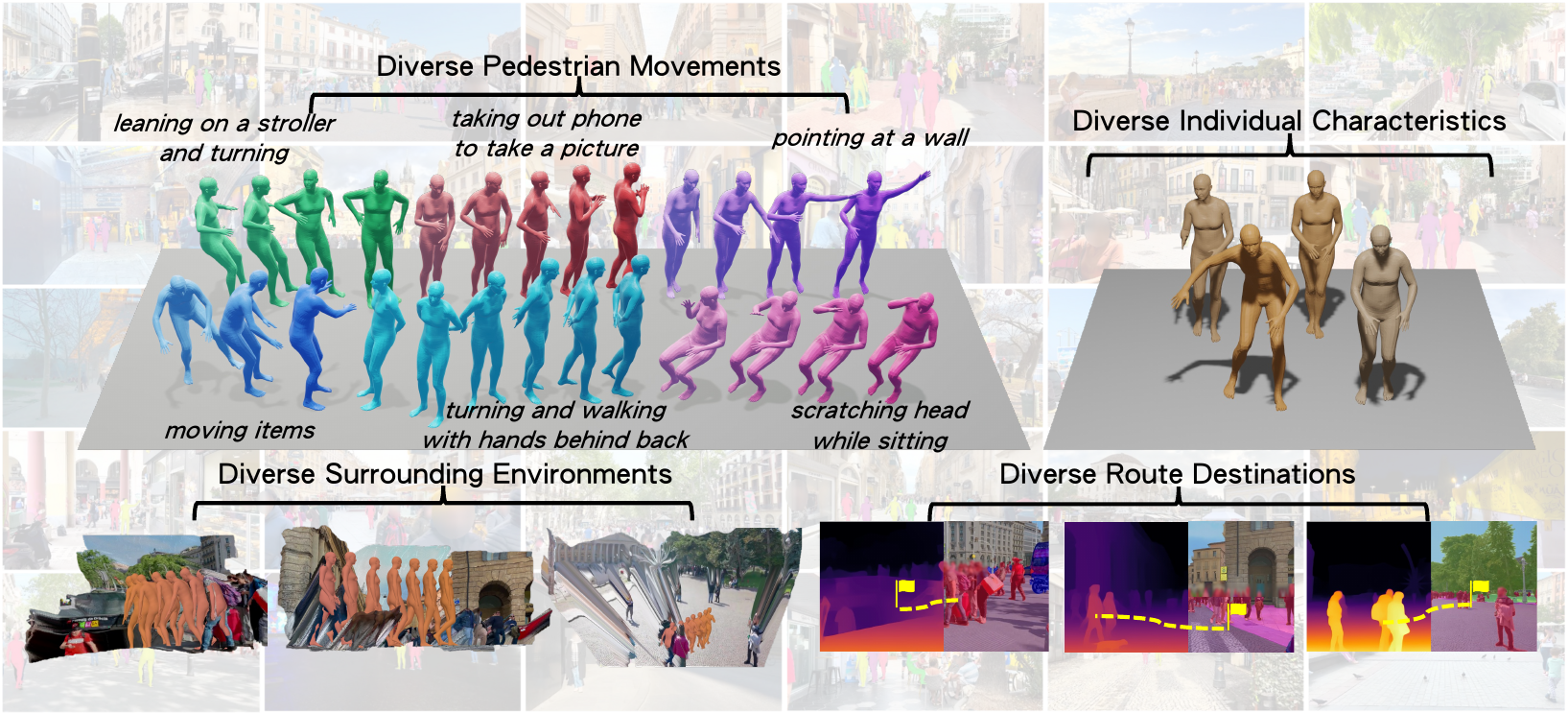}
    \caption{\textbf{Samples in the CityWalker dataset.} Top Left: The diverse pedestrian movements. Top Right: The diverse body shapes of the pedestrians.
    Bottom Left: The diverse surrounding environments from depth-unprojected images and the 4D pedestrian movement labels.  
    Bottom Right: The diverse route destinations shown on the depth labels and semantic maps of the scene.
    The background showcases pedestrians in bustling cities from where we construct the CityWalker dataset.
    }
    \label{fig:dataset}
        
\end{figure}

The CityWalkers dataset is collected from YouTube's city tour videos. Our data source consists of high-quality web videos of walking in cities worldwide posted by content creators on YouTube, such as the YouTube channel POPtravel~\citep{youtube_poptravel}. We manually picked videos so they incorporate a wide spectrum of urban public places in different regions and cultures with various scene contexts and real-world pedestrian movements. We label each video's scene attributes, including weather, location, crowd density, and time of the day, using an off-the-shelf VLM~\citep{chen2023internvl}. We then detect~\citep{Jocher_Ultralytics_YOLO_2023}  and track~\citep{cheng2023tracking} each pedestrian in the video, and label the pedestrian movements as SMPL parameters and scene context as depth and semantic segmentation maps. These labels contain the crucial context factors for learning pedestrian movements. In total, CityWalker contains 30.8 hours of high-quality videos, including 120,914 pedestrians and 16,215 scenes across 227 cities and 49 countries, making it the most diverse human motion dataset regarding scene context and human subjects. As shown in Fig.~\ref{fig:dataset}, CityWalkers contains a variety of pedestrian movements, scene contexts, pedestrian body shapes, and route destinations.

\subsection{Data Processing and Autolabeling}
\label{dataset_label}
We adopt WHAM~\citep{shin2023wham} to recover 4D pedestrian movement pseudo-labels. Given tracked pedestrians and their body key points, WHAM outputs their global human motion with SMPL~\citep{loper2023smpl} parameters $\{\boldsymbol{t}_{t}, \boldsymbol{\phi}_{t}, \boldsymbol{\theta}_{t}, \boldsymbol{\beta}\}$ at every timestep $t$. $\boldsymbol{t}_{t}\in \mathbb{R}^3$, $\boldsymbol{\phi}_{t}\in SO(3)$ are the root translation and orientation of the SMPL model, and $\boldsymbol{\theta}_{t}\in SO(3)^{23}$, $\boldsymbol{\beta} \in \mathbb{R}^{10}$ are the body pose and shape parameters. We further use the state-of-the-art monocular depth estimation model ZoeDepth~\citep{bhat2023zoedepth} and semantic segmentation model SegFormer~\citep{xie2021segformer} to automatically label scene depth and semantics. To filter low-quality pedestrian movement labels brought by model noise,  we threshold the detection confidence score, human bounding box size, and keypoint confidence scores. We also discard the occluded motion by thresholding the number of human key points that are within the 2D human segmentation mask. To further clean the labels and curate a more accurate evaluation dataset, we manually check the label quality and adjust the labels, such as 2D key points, if necessary. We want to stress that though we have tried our best to improve the quality of pseudo labels by using state-of-the-art models and filtering wrong predictions, label noise from web videos is still inevitable. We show in the experiments that learning from the noisy labels still benefits pedestrian movement generation. We also apply techniques in our generation model to further mitigate the effect of label noise. A benchmark result of the accuracy of our data autolabeling pipeline is provided in the Appendix.

The final label includes future pedestrian movements and their corresponding context factors. $\{\boldsymbol{x}=\{(\boldsymbol{t}_{t}, \boldsymbol{\phi}_{t}, \boldsymbol{\theta}_{t}) \}, \boldsymbol{y}=[\mathcal{I}, \mathcal{I}^d$, $\mathcal{I}^s,  \boldsymbol{\beta}, \boldsymbol{t}_1, \boldsymbol{t}_T]\}$, where $\boldsymbol{x}$ is the pedestrian movement label and $\boldsymbol{y}$ is the context label. $\mathcal{I}, \mathcal{I}^d$, $\mathcal{I}^s$ are images,  depth labels, and segmentation maps of the scene,  $\boldsymbol{\beta}$ represents the pedestrian's personal characteristics, and $\boldsymbol{t}_1, \boldsymbol{t}_T$ are the starting and goal positions. 

\paragraph{Mitigating the data privacy and ethics issues.} We are fully aware of the potential privacy and ethics issues when using street-view videos, and we take it as the highest priority. We have taken several measures to mitigate that: 1) We follow standard protocols used by street datasets like Waymo~\citep{sun2020scalability} to ensure complete anonymization of identifiable features. 2) We will provide clear and transparent terms of use for the dataset, which outline the ethical guidelines, usage restrictions, and legal obligations that users must comply with. 3) Our dataset is overseen by an independent ethics review board with security protocols to protect the dataset from unauthorized access and misuse. Please refer to the Appendix for the details.

\section{Generating Context-Aware Pedestrian Movements from Noisy Labels}
\label{sec:method}
\begin{figure}[t!]
    \centering
    \includegraphics[width=1.0\linewidth]{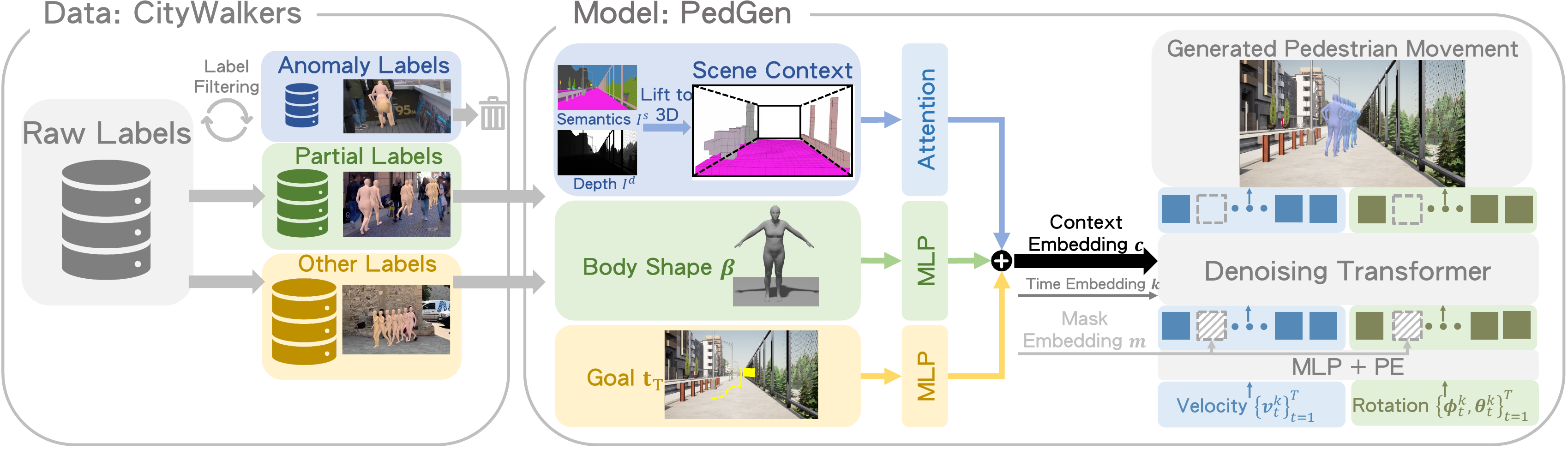}
    \caption{\textbf{Our method}. We discard the anomaly labels with an iterative automatic label filtering procedure and add the partial labels to training data. We then train PedGen with a Context Encoder to represent crucial context factors. The scene context is obtained by lifting the 2D depth and semantic labels to the 3D space and converting them into a local voxel representation. The encoded scene context is combined with other context factors, including the body shape and the goal to get the context embedding $\boldsymbol{c}$. The context embedding $\boldsymbol{c}$ and the timestep embedding $\boldsymbol{k}$ are then used to guide the Denoising Transformer to predict the clean motion from the noised one. We use a learnable motion mask embedding $\boldsymbol{m}$ to address the partial labels during training.
    }
    \label{fig:method}
        
\end{figure}

This section introduces our method, PedGen, to address the label noise and incorporate the context factors from the CityWalkers dataset.
PedGen is a diffusion-based generative model and the first method for the new task of context-aware pedestrian movement generation. An overview of PedGen is shown in Fig.~\ref{fig:method}. Sec.~\ref{task_definition} defines the task of context-aware pedestrian movement generation. We introduce the overall architecture of PedGen in Sec.~\ref{pedgen_architecture}. Sec.~\ref{sec:handle_labels} introduces two simple and effective strategies to deal with the low-quality and incomplete labels from web videos, respectively. As it is crucial to model the context factors, including the surrounding environments, the pedestrian's characteristics, and the goal points during generation, we design a novel Context Encoder in Sec.~\ref{sec:context_encoder}.  

\subsection{Task Definition}
\label{task_definition}

We define the task of \textit{context-aware} pedestrian movement generation as follows: Given the initial 3D position $\boldsymbol{t}_1=[x_1,y_1,z_1]$ of a pedestrian and the 2D image of an urban scene, our goal is to generate the pedestrian's future movements $\boldsymbol{x}=[\boldsymbol{t}_{t}, \boldsymbol{\phi}_{t}, \boldsymbol{\theta}_{t}]_{t=1}^T$. The movement at each timestep is represented as the SMPL root translation $\boldsymbol{t}_{t}$, root orientation $\boldsymbol{\phi}_{t}$, and body pose $\boldsymbol{\theta}_{t}$. The following context factors are also provided to help generation: 1). The urban scene context represented as the  2D image $\mathcal{I}$, semantic mask  $\mathcal{I}^s$, and depth label $\mathcal{I}^d$. 2). The SMPL human shape parameter $\boldsymbol{\beta}$, which is a latent representation that can indicate a pedestrian's characteristics, such as height, weight, and body shape. 3). The 3D goal position $\boldsymbol{t}_T=[x_T, y_T, z_T]$ as the pedestrian's route destination  in the scene.

\subsection{PedGen Model}
\label{pedgen_architecture}
PedGen follows the conditional diffusion framework~\citep{ho2020denoising, ho2022classifier}. Given a sampled movement $\boldsymbol{x}$ from the dataset, the forward diffusion is a Markov noising process $\{\boldsymbol{x}^k\}_{k=0}^K$ defined from the noise scheduling parameter $\{\alpha^k\}_{k=1}^K$. The goal is to train a reverse denoising model $\hat{\boldsymbol{x}} = F(\hat{\boldsymbol{x}}^k, k, \boldsymbol{c})$ that predicts the sampled movement $\boldsymbol{x}$ given the noisy movement  $\hat{\boldsymbol{x}}^k$, the diffusion timestep $k$, and the condition factor $\boldsymbol{c}$. Our denoising model architecture follows state-of-the-art transformer-based human motion generation models~\citep{tevet2022human, jiang2024scaling, chen2023humanmac}, where movements at different timesteps are encoded as separate tokens in the transformer.
We represent  $\boldsymbol{x}$ as velocity  $\boldsymbol{v}_t = \boldsymbol{t}_t - \boldsymbol{t}_{t-1}$ and rotation $\boldsymbol{\phi}_t, \boldsymbol{\theta}_t$ with 6D rotation~\citep{zhou2019continuity} representation.
Different from existing approaches that encode all parts of the motion into a single token, we find it helpful to treat the velocity and the rotation of the motion as different tokens in the transformer, as they have different representations and scales in the data. 
Our loss function for training the denoising model $F$ is as follows:
\begin{equation}
\mathcal{L} (\boldsymbol{x}, \hat{\boldsymbol{x}}) = \mathbb{E}_{k \in [1, K], (\boldsymbol{x}, \boldsymbol{c}) \in \mathcal{D}}[w_{\mathrm{rec}} \mathcal{L}_{\mathrm{rec}} + w_{\mathrm{traj}}  \mathcal{L}_{\mathrm{traj}} + w_{\mathrm{geo}}\mathcal{L}_{\mathrm{geo}}],
\end{equation}
where $\mathcal{D}$ is the training dataset. $\mathcal{L}_{\mathrm{rec}} = \lVert \boldsymbol{x} - \hat{\boldsymbol{x}} \rVert_2^2 $ is the standard diffusion reconstruction loss. $\mathcal{L}_{\mathrm{traj}} = 
\sum_{t=1}^T \lVert \boldsymbol{t}_t - \hat{\boldsymbol{t}}_t - \boldsymbol{t}_0  \rVert_1$ is the trajectory loss on the reconstructed global translation $\hat{\boldsymbol{t}}_t = \sum_{t=1}^T  \hat{\boldsymbol{v}}_t$, and $\mathcal{L}_{\mathrm{geo}} = \sum_{t=1}^T  \lVert \mathrm{FK}(\boldsymbol{\theta}_t) - \mathrm{FK}({\hat{\boldsymbol{\theta}}_t}) \rVert_2$ is the geometric loss~\citep{tevet2022human} that computes the joint position error with forward kinematics on the SMPL body pose. In ablation experiments, we will show the effectiveness of separating velocity and rotation tokens, as well as adding the trajectory and geometry losses.

 During inference,  we start from the $k=K$ step by sampling $\hat{\boldsymbol{x}}^K$ from the standard Gaussian distribution. We then pass it to the denoising model to predict the clean motion $\hat{\boldsymbol{x}}$ and add noise again to obtain the noised sample $\hat{\boldsymbol{x}}^{K-1}$. This process is repeated $K$ times to get the final generated pedestrian movement $\hat{\boldsymbol{x}}$. 
Our model can also facilitate the efficient generation of infinitely long movements by concatenating the short motion intervals. Please refer to the appendix for more details.

\subsection{Addressing Noisy Labels}
\label{sec:handle_labels}
Pseudo-labels from pre-trained predictors have inevitable noise. Unlike existing human motion generation approaches that focus on datasets with clean MoCap data, our PedGen model aims to address noisy labels from web videos. There are two sources of label noise in CityWalkers. The first is the inherent compound noise from the data and models used for pseudo-labeling, leading to low-quality labels. As it is difficult to examine the label quality of 4D human motion from only 2D videos, many anomaly labels remain in the dataset even after rule-based filtering and manual checks. We propose to automatically identify these low-quality labels using reconstruction-based unsupervised anomaly detection techniques~\citep{livernoche2023diffusion, wolleb2022diffusion}. Specifically, we first train a PedGen model without context on the training data of CityWalkers. We then partially add noise with half of the diffusion steps $K/2$ for each sample and denoise it using the trained model. We use the reconstruction error between the original and the denoised sample as a metric for anomaly labels and filter labels with errors greater than a certain threshold. We then iterate the above process by re-training the model with the remaining labels and filtering based on the new reconstruction error.

Another source of label noise is the discontinuous or lost tracks of pedestrians due to occlusions and missed detections. As a result, more than half of the labels are incomplete and only annotated on a subset of frames. These labels provide partial supervision that can benefit learning more diverse context-aware pedestrian movements. To train with these partial labels, we replace the missing timesteps for these partial labels with a learnable motion mask embedding $\boldsymbol{m}$ to the denoising transformer. The loss is only computed with the labels at the available timesteps.

\subsection{Context Encoder}
\label{sec:context_encoder}
Our context encoder module outputs a condition embedding $\boldsymbol{c}$ from the provided context factors.  To encode the scene context, we first unproject the 2D depth label $\mathcal{I}^d$ and semantic map $\mathcal{I}^s$ into a 3D point cloud $\mathcal{P} = \{\mathbf{p}=[p_x, p_y, p_z, p_c] \}$, where $p_c$ is the semantic label.
Next, we extract points within a local neighborhood of the starting location  $\boldsymbol{t}_1$, resulting in the local point cloud $\mathcal{P}_{\mathrm{local}}=\{\mathbf{p} \in \mathcal{P} \mid \lVert p_x- x_1 \rVert < \Delta_x, \lVert p_y-y_1 \rVert < \Delta_y, \lVert p_z-z_1\rVert < \Delta z\}$ and voxelize it into a 3D grid. The class label of each voxel is either empty or determined by majority voting of the points within the voxel. The voxel is then processed with a single cross-attention layer to get the scene context embedding. The scene encoding is then added with embeddings from other conditions, including the human shape $\boldsymbol{\beta}$ and the goal position $\boldsymbol{t}_T$, to get the final context embedding $\boldsymbol{c}$.

As web videos only contain 2D images, a natural idea is to directly encode the scene context using the 2D image feature extractors (e.g., Dino-V2~\citep{oquab2023dinov2}). We find that our proposed context encoder can encode the scene context more effectively than encoding the scene context in 2D. Since pedestrian movement is represented in the 3D space and generating it requires 3D information about the scene, it is easier to reason about the surrounding environment using a 3D representation than 2D image features. Furthermore, it is hard to disentangle the pedestrian and its surrounding context from the 2D image features. As a result, the model may exploit the ego pedestrian instead of the scene context to generate future movements. Our context encoder could address this issue by unprojecting the pixels into 3D point clouds and discarding point clouds belonging to the ego pedestrian. In our ablation experiments, we show the importance of using the 3D scene representation and appending the semantic labels in encoding the scene context.

\section{Experiments}
\label{sec:exp}

\begin{figure}[t!]
    \centering
    \includegraphics[width=0.9\linewidth]{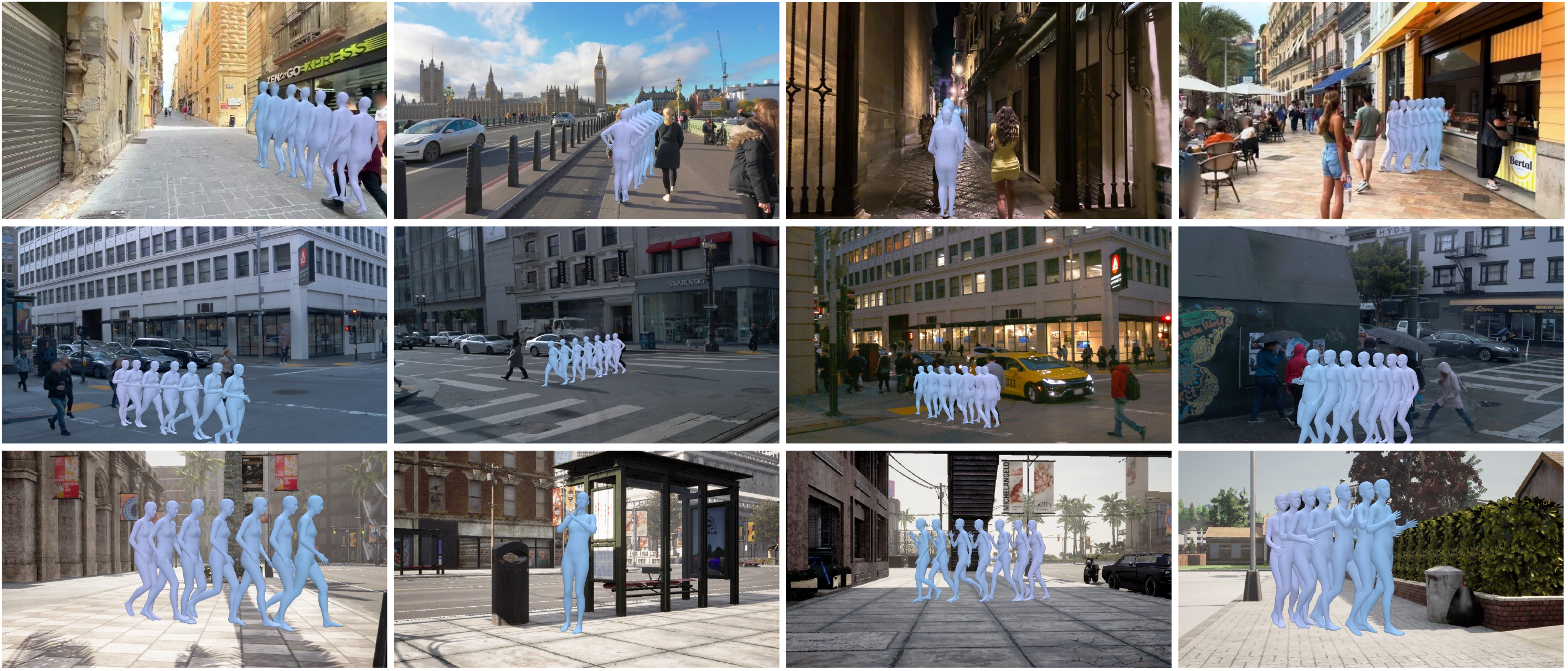}
    \caption{\textbf{Visualizations of the generated pedestrian movements.} The top row shows results in real scenes from the CityWalkers dataset, the middle row shows results in the real-world Waymo test set, and the bottom row shows results in simulated scenes from the CARLA test set.   
    }
    \label{fig:qualitative}
   
\end{figure}

We compare the performance of PedGen to the other baselines on the real-world Citywalkers and Waymo datasets and in simulated CARLA environments in Sec.~\ref{eval_real_world}. Sec.~\ref{eval_simulation} shows experiments on the effect of training with noisy labels and different context factors. Sec.~\ref{ablation} demonstrates the ablation study of our data and model. The diverse pedestrian movements generated by PedGen are shown in Fig.~\ref{fig:qualitative}. Sec.~\ref{sec:more_qualitative} demonstrates additional qualitative results in real-world applications and we provide video qualitative results and additional experimental details in the Appendix.

\paragraph{Datasets.}
We train PedGen on the proposed CityWalkers dataset. For each pedestrian movement trajectory, we sample the initial timestep at an interval of 30 frames and keep at most the future 60 frames (2 seconds) as the ground truth movement.  The training set has 104,192    samples, including 53,405 partial labels that have at least 30 frames of annotation. The validation set has 13,039 samples and only contains complete labels.  We split the validation set to contain novel scenes with completely different locations and human subjects never seen in the training set.

To verify the performance of PedGen on real-world pedestrian movement prediction with ground truth labels,  we use the Waymo open dataset~\citep{sun2020scalability} with human-annotated 3D human keypoint labels, which is only used for testing. Waymo is critical as it is the largest urban dataset that captures diverse pedestrian motions with sparse 3D keypoint labels at 10 Hz. In total, we selected 80 test samples that (1) spans 2 seconds and (2) includes at least 6 sparse human keypoint labels.

We also collect an additional test set in the CARLA Simulator~\citep{dosovitskiy2017carla} to demonstrate the application of PedGen in simulation. We sample initial locations, goal locations, and camera view angles in different maps and render the ground truth images, depth labels, and segmentation maps in simulation. We manually check that all the sampled locations are valid and that all the camera views are not occluded by obstacles, resulting in a simulated test set with 262 diverse samples.

\paragraph{Evaluation metrics.}
For experiments on the CityWalkers and Waymo dataset, we compare the generated pedestrian movement with the ground truth, and follow the metrics used in human motion prediction~\citep{chen2023humanmac} to compute Average Displacement Error (ADE) and Final Displacement Error (FDE) on the realism of the predicted motion.  We generate 50 movements for each data sample and report both minimum and average ADEs (mADE, aADE) and FDEs (mFDE, aFDE) among all samples.
For experiments on the simulated test set, we evaluate the context awareness and the physical plausibility of the generated motion. Our metrics include the Collision Rate (CR), which measures the ratio of the generated movement that collides with other objects in the environment, and the Foot Floating Rate (FFR), which measures the ratio of the generated movement whose feet are either floating or penetrating with the ground greater than a given threshold (20cm).

\paragraph{Baselines.} We compare our method with several recent human motion diffusion models, which have different input conditions and motion representations.
Note that since context-aware pedestrian movement generation is a newly defined task, no models could directly support it; we thus made minimum adjustments to make them compatible with our problem setting.
MDM~\citep{tevet2022human} uses texts or action labels as the condition and proposes several geometric losses to improve motion quality. We did not use the foot contact loss proposed in their method as CityWalkers does not provide ground-truth foot contact labels.
HumanMac~\citep{chen2023humanmac} conditions the future motion generation on the history motion sequences, which uses the DCT transform to ensure smooth motion generation.
TRUMANS~\citep{jiang2024scaling} uses the initial human pose, the 2D BEV goal position, the indoor scene context, and the frame-wise action labels as the input condition. We modify TRUMANS so it is compatible with our scene context and goal point conditions.

\subsection{Comparison to baselines}
\label{eval_real_world}
    
\begin{table}[t] 
\centering
\scriptsize
\setlength\tabcolsep{1.5pt}
\caption{\textbf{Comparison to baselines.} We consider two cases where the model is given the goal condition or not. We evaluate on the validation set of CityWalkers, the real-world Waymo test set, and the simulated CARLA test set. We evaluate our model (PedGen) compared with the other baselines.}
        \label{tab:main_benchmark}
\begin{tabular}{@{}cccccc|cccc|cc@{}}
\toprule
    \multirow{2}{*}{\begin{tabular}[c]{@{}c@{}}\textbf{Goal}\\ \textbf{Condition}\end{tabular}}& \multirow{2}{*}{\textbf{Method}} & \multicolumn{4}{c}{\textbf{CityWalkers} }  & \multicolumn{4}{c}{\textbf{Waymo}  }   & \multicolumn{2}{c}{\textbf{CARLA}  }                    \\  \cmidrule(l){3-12}  
    &   & mADE $\downarrow$       & aADE $\downarrow$ & mFDE $\downarrow$ &  aFDE $\downarrow$  & mADE $\downarrow$       & aADE $\downarrow$ & mFDE $\downarrow$ &  aFDE $\downarrow$ & CR $\downarrow$ &  FFR $\downarrow$  \\ \midrule
\xmark & HumanMAC              &      1.31        &  4.67              &  1.86     & 8.65 & 3.19 & 5.29 & 5.61 & 10.36 & 2.5\% & 10.2\% \\
 \xmark & MDM                &      1.33       &  4.55              &  1.93     & 8.41 & 3.03&  5.35 &  5.66 &  10.60 & 2.1\% & 3.2\% \\

     \xmark &  PedGen                  &   \textbf{1.13}         &   \textbf{4.08}             &      \textbf{1.61}      &    \textbf{7.56}  & \textbf{2.90}& \textbf{5.15}& \textbf{5.52}& \textbf{10.11}& \textbf{1.6\%} & \textbf{2.6\%}     
      
      \\ \midrule

       \cmark &      TRUMANS               &      0.73       &  1.26              &  0.56     & 1.13 & 2.01 & 2.37 & 1.41 & 1.94 &0.6\% & 0.6\% \\
     \cmark & PedGen                             &    \textbf{0.59}        &   \textbf{1.08}                &     \textbf{0.46}       &  \textbf{0.99}  & \textbf{1.91} & \textbf{2.18}& \textbf{0.78}& \textbf{1.04} & \textbf{0.0\%} & \textbf{0.0\%}        \\    \bottomrule
\end{tabular}
\end{table}
We compare PedGen with the other baselines on the validation set of CityWalkers, the real-world Waymo test set, and the simulated CARLA test set. We separately evaluate whether the goal condition is given, as it is a deterministic factor for pedestrian movement. The benchmarking results are shown in Tab.~\ref{tab:main_benchmark}. It can be observed that PedGen outperforms other baseline models by a clear margin on CityWalkers. PedGen also achieves the best zero-shot generalization ability on Waymo and CARLA test sets, further showing the superiority of PedGen to the other baseline.

\begin{figure}[t!]
    \centering
    \includegraphics[width=0.9\linewidth]{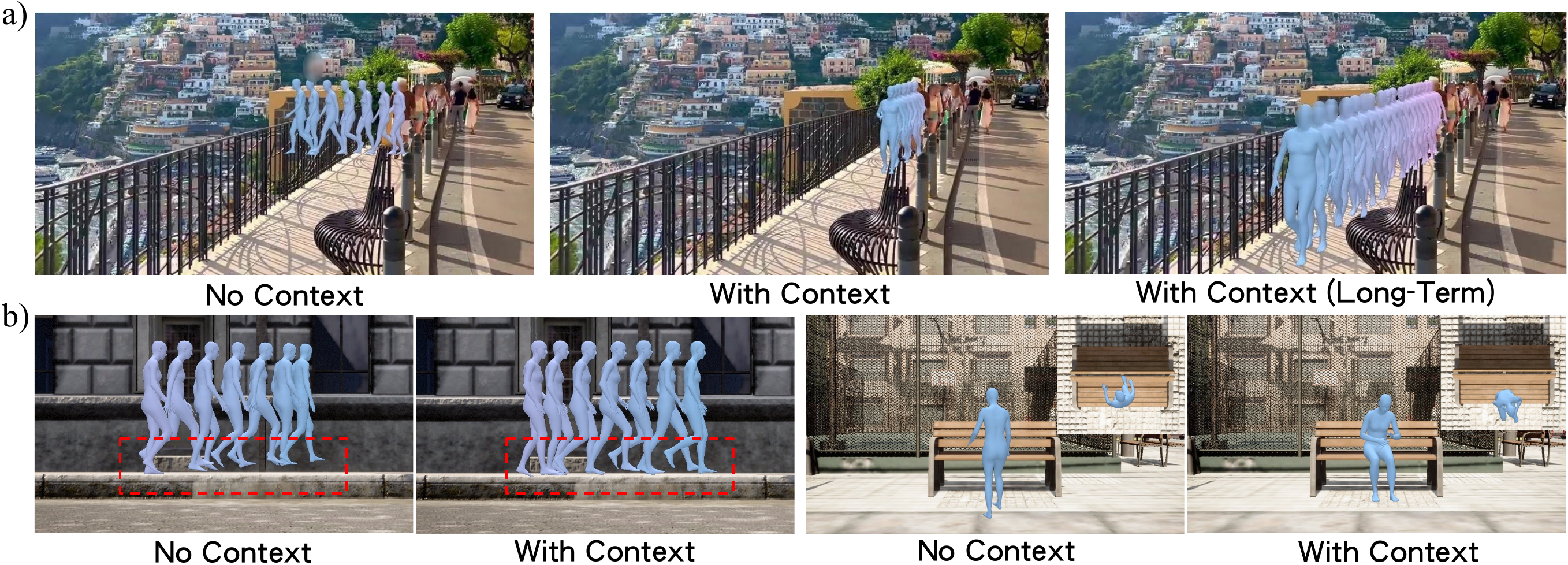}
    \caption{\textbf{Qualitative comparison of training with context factors.} We compare the generated movements of PedGen trained with or without context factors in real-world environments (a) and in simulation (b).
    }
    \label{fig:carla_vis}

\end{figure}
\begin{table}[htbp]
\scriptsize

\setlength\tabcolsep{1.0pt}
\caption{\textbf{Results with noisy labels and context factors.}  We experiment on the CityWalkers validation set and study the effect of training with noisy labels (a) and different context factors (b).}

\begin{subtable}[t]{0.45\textwidth}
\scriptsize
\centering
\caption{\textbf{Evaluation of training with noisy labels.} We evaluate PedGen with no context trained with or without anomaly and partial labels.}
     \label{tab:main_a}
\begin{tabular}{@{}cccccc@{}}
\toprule

\multirow{2}{*}{\begin{tabular}[c]{@{}c@{}}\textbf{Anomaly} \\ \textbf{Labels}\end{tabular}} & \multirow{2}{*}{\begin{tabular}[c]{@{}c@{}}\textbf{Partial} \\ \textbf{Labels}\end{tabular}} & \multicolumn{4}{c}{\textbf{Metric}}   \\ \cmidrule(l){3-6} 
 &                                                                               & mADE $\downarrow$   & aADE $\downarrow$   & mFDE $\downarrow$  & aFDE $\downarrow$    \\ \midrule
\cmark                                                                           & \xmark                                                                            &     1.17   &  4.45      &    1.64   & 8.31         \\
\xmark                                                                           & \xmark                                                                            &    1.13     &  4.32      &  \textbf{1.60}      &    8.09      \\
\cmark                                                                           & \cmark                                                                            & 1.15       &    4.19     & 1.62       &    7.79      \\
\xmark                                                                           & \cmark                                                                          &   \textbf{1.13}     &   \textbf{4.08}     &   1.61    &  \textbf{7.56}           \\
  \bottomrule
\end{tabular}
\end{subtable}
                   \, \, \, \,
                   \begin{subtable}[t]{0.5\textwidth}
\centering
\caption{\textbf{Evaluation of the context factors.} We evaluate PedGen conditioned on each context factor, including the surrounding environment (scene), the pedestrian's own characteristics (human), and the goal points (goal).}
        \label{tab:main_b}
\begin{tabular}{@{}cccccc@{}}
\toprule
 \multirow{2}{*}{\textbf{Idx}}  &  \multirow{2}{*}{\begin{tabular}[c]{@{}c@{}}\textbf{Context}\\ \textbf{Factor}\end{tabular}}& \multicolumn{4}{c}{\textbf{Metric} }                    \\  \cmidrule(l){3-6}  
     &   & mADE $\downarrow$       & aADE $\downarrow$ & mFDE $\downarrow$ &  aFDE $\downarrow$    \\ \midrule

1& No            &      1.13        &  4.08              &       1.61     &     7.56       \\
      
         
    2 &   Scene            &     1.11      &     3.75            &       1.55     &     6.92       \\

      3 &     Human                   &   1.09         &   3.24             &      1.61      &    5.95    \\  
  4&     Scene + Human                   &   \textbf{1.00}         &   \textbf{3.12}             &      \textbf{1.43}      &    \textbf{5.65}      
      
      \\ \midrule
            
    5&     Goal             &   0.60         &    1.09             & 0.47           &      1.00      \\
    
     6 &  Scene + Goal                              &    0.59           &  1.08                  &      0.46      &    0.99        \\
  7 &     Human + Goal                           &         0.55   & 1.01                    &    0.43        &  0.93          \\8 & Scene + Human + Goal                              &    \textbf{0.54}        &   \textbf{0.96}                &     \textbf{0.43}       &  \textbf{0.91}          \\    \bottomrule
\end{tabular}
   
\end{subtable}

               
       
\end{table}

\subsection{Effect of  Noisy Labels and Context Factors}
\label{eval_simulation}
Table~\ref{tab:main_a} demonstrates the effectiveness of PedGen in addressing noisy labels. We can see that removing the anomaly labels with the proposed automatic label filtering can help generate more realistic pedestrian movements and improve aADE by 2.9\%. Moreover, adding the partial labels as additional training data can improve the aADE by 5.8\%, highlighting the value of partial labels in web videos. Combining both strategies for noisy labels leads to the best performance of a 4.08 aADE. 

We further examine the effect of each context factor on the pedestrian movement generation performance in  Tab.~\ref{tab:main_b}. The comparisons between ``setting 1 vs. 2'' and between ``setting 1 vs. 3'' show that adding the surrounding environment and the pedestrian's own characteristics as conditions are helpful to pedestrian movement generation and can reduce the aADE by $6.7\%$ and $19.4\%$, respectively. Setting 4 demonstrates that incorporating both context factors achieves even better results. From settings 1 and 5, we find that the goal points are the most important context factor for the final performance and can significantly reduce aADE by $72.9\%$. The comparison between ``setting 5 vs. 6'' and between ``setting 5 vs. 7'' further proves that the other two context factors (scene and human) can also help movement generation after the goal is provided. Finally, setting 8 demonstrates that using all three context factors leads to the smallest generation errors.

Some qualitative comparison results in real-world environments are shown in Fig.~\ref{fig:carla_vis} (a). We can observe that PedGen trained without context factors generate arbitrary movements that walk off the sidewalk. By conditioning the generation on the context factors, the movement becomes context-aware, and the model can further generate long-term pedestrian walking behaviors on the sidewalk. Visualization results in simulation are shown in Fig.~\ref{fig:carla_vis} (b). From the left two figures, we can see that without the context factors, the generated movements have the feet floating from the ground, while adding context factors fix this issue. From the right two figures, we see that PedGen trained without context generated a walking pose that is in collision with the bench, while it successfully generated a sitting pose on the bench after considering the context factors.

\subsection{Ablation Study}
\label{ablation}

\noindent
\textbf{Ablation on training data of PedGen.}
Table~\ref{tab:ab_a} shows an ablation study on the training data of PedGen. To demonstrate the effectiveness and necessity of using web videos for pedestrian movement generation, we train PedGen on SLOPER4D~\citep{Dai_2023_CVPR}. SLOPER4D is one of the largest outdoor 4D human datasets annotated from LiDAR point clouds. Still, its scale and diversity in human subjects and scene contexts are much less than CityWalkers, and its motion is captured from human actors instead of real-world pedestrians. We can observe that training on CityWalkers leads to significant performance improvement compared to training on SLOPER4D, even though CityWalkers is annotated with pseudo-labels. Notably, training with the human context on SLOPER4D only achieves 3.82 mADE, as the model can easily overfit the 12 human subjects in the dataset and results in degraded generalization performance. On the contrary, with more training data of CityWalkers, models conditioned on different context factors all perform better than those trained with less data, highlighting the value of capturing large-scale pedestrian movement data using web videos. 

\noindent
\textbf{Ablation on model components of PedGen.}
Table~\ref{tab:ab_b} shows ablations on key model components of PedGen. \textit{-traj/geo} means the model is trained only with the diffusion reconstruction loss without the trajectory and geometry losses.
\textit{-sep. token} means the motion at each timestep is represented as a single token in the transformer instead of separate tokens for the rotation and velocity. \textit{-3D rep.} means the scene context is encoded with 2D feature using a depth anything~\citep{depthanything} pre-trained image backbone DINOv2~\citep{oquab2023dinov2}. \textit{-semantic} means the 3D voxel only encodes the occupancy without semantic labels. The results show the importance of using 3D scene context with semantic labels for pedestrian movement generation, the effectiveness of separating velocity and rotation tokens, and adding the geometry and trajectory losses in the motion diffusion model.

\subsection{More Qualitative Results}
\label{sec:more_qualitative}

\begin{figure}[t!]
    \centering
    \includegraphics[width=0.9\linewidth]{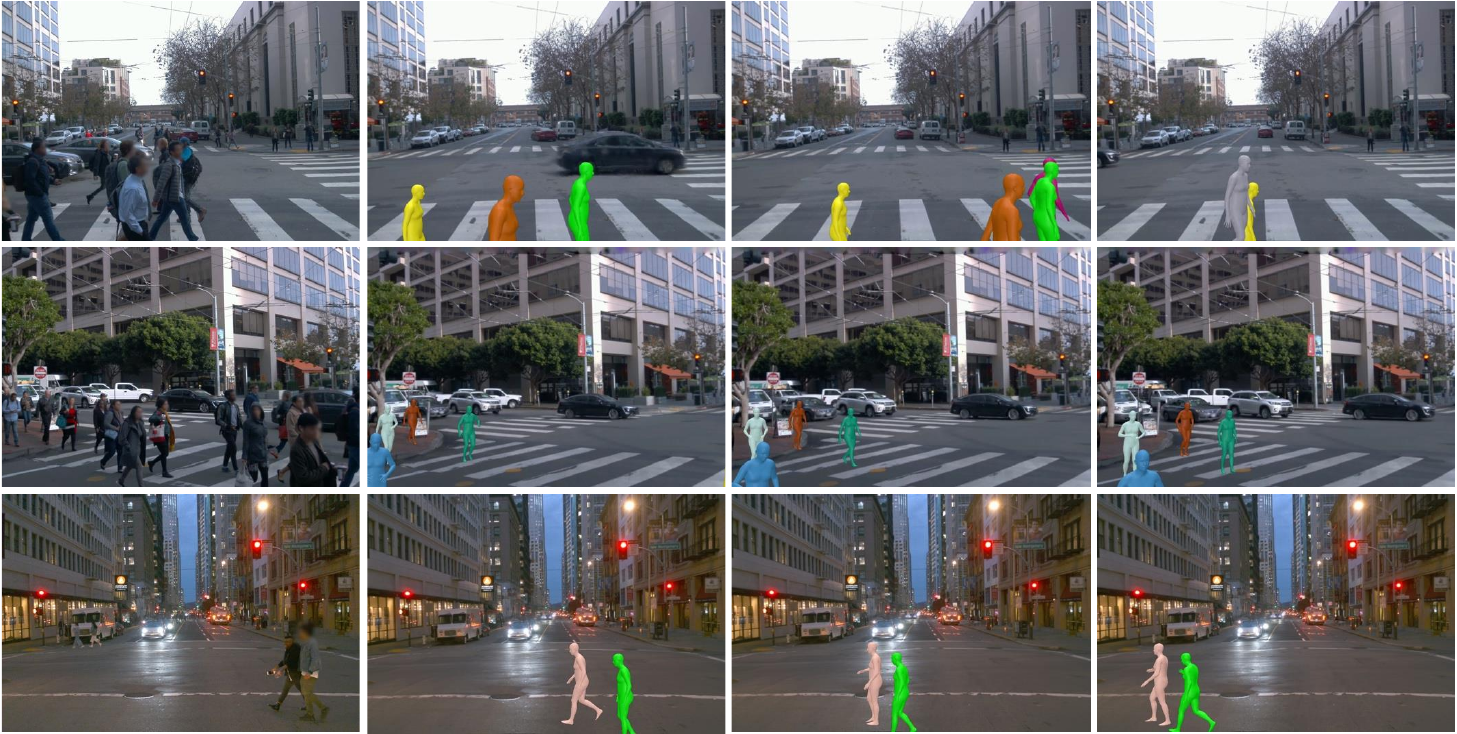}
    \caption{\textbf{Pedestrian movement prediction in Waymo.} We predict long-term pedestrian movements using PedGen.   
    }
    \label{fig:demo_waymo}
   
\end{figure}

\begin{figure}[t]
    \centering
    \includegraphics[width=0.9\linewidth]{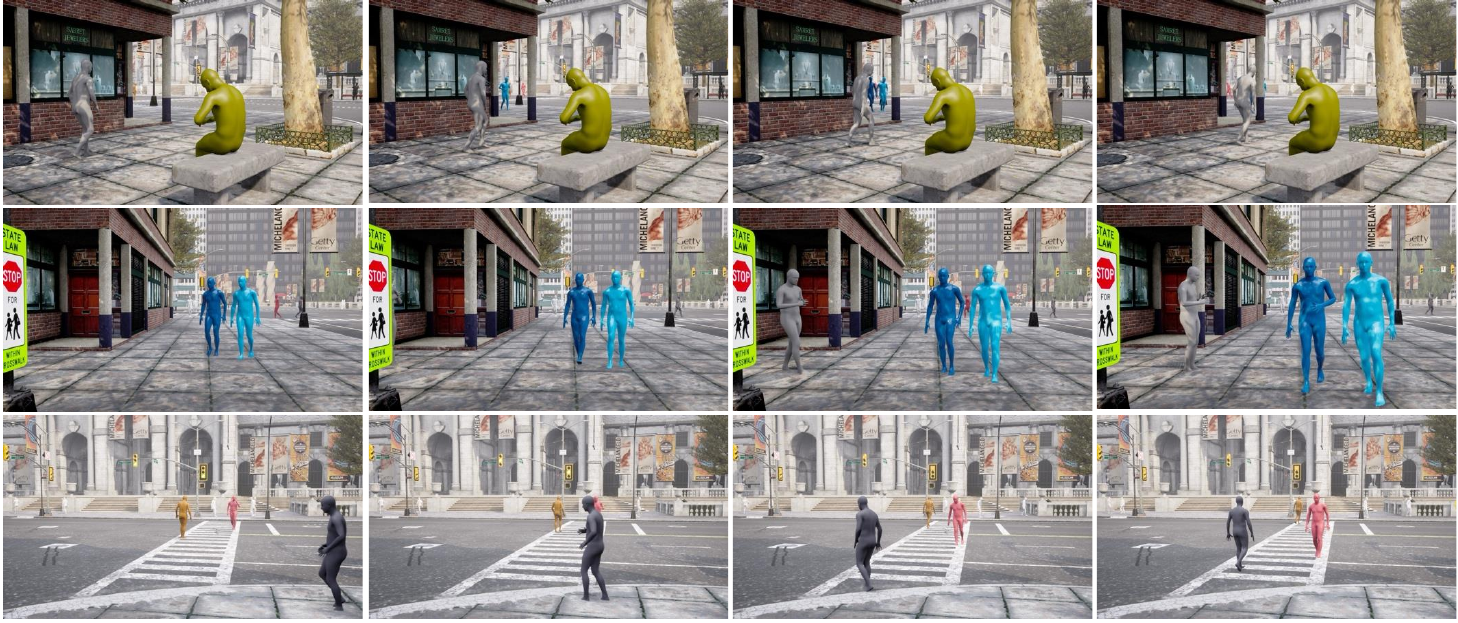}
    \caption{\textbf{Populating
urban scenes in CARLA.} We generate multi-pedestrian movements using PedGen.   
    }
    \label{fig:demo_carla}
   
\end{figure}

Fig.\ref{fig:demo_waymo}  shows the application of PedGen in real-world pedestrian movement prediction by generating diverse long-term pedestrian movements in a 3D Gaussian-Splatting reconstructed~\cite{chen2024omnire} scenario in Waymo. Fig.\ref{fig:demo_carla}  shows the application of PedGen in simulated environments by populating urban scenes in CARLA with realistic multi-pedestrian movements. The results further demonstrate the practicality of PedGen in real-world applications.

\begin{table}[t!]
\scriptsize
\setlength\tabcolsep{1.5pt}
\caption{\textbf{Ablation experiment results.} We ablate on the training data of PedGen (a) and the PedGen model’s key components (b) on the CityWalkers validation set.  }
 
 \begin{subtable}[t]{0.48\textwidth}
    \caption{Ablation on training data of PedGen. }
        \label{tab:ab_a}
 
\begin{tabular}{cccccc@{}}
\toprule
 \multirow{2}{*}{\begin{tabular}[c]{@{}c@{}}\textbf{Context}\\ \textbf{Factor}\end{tabular}}    & \multirow{2}{*}{\begin{tabular}[c]{@{}c@{}}\textbf{Training}\\ \textbf{Data}\end{tabular}} & \multicolumn{4}{c}{\textbf{Metric} }                                                                      \\ \cmidrule(l){3-6}  
                 &             & mADE $\downarrow$       & aADE $\downarrow$ & mFDE $\downarrow$ &  aFDE $\downarrow$           \\ \midrule
\multirow{3}{*}{ No}  
 &SLOPER4D                                                                     &    1.61                   &   6.04                                &    2.42                  &        11.65   \\
 &CityWalkers($50\%$)                                                                     &       1.16               &    4.19                              &      1.66               &  7.76        \\
&  CityWalkers($100\%$)                                                    &     \textbf{1.13}                 &            \textbf{4.08}             &  \textbf{1.61}                    &   \textbf{7.56}                   \\ \midrule
\multirow{3}{*}{ Scene}  
 &SLOPER4D                                                                     &   1.47                    &   6.45                                &     2.19                 &   12.50        \\
&  CityWalkers($50\%$)                                                                     &    1.14                  &          4.02                      &    1.57                   &     7.47   \\
& CityWalkers($100\%$)              &  \textbf{1.11}                    &       \textbf{3.75}                  &  \textbf{1.55}                    & \textbf{6.92}                     \\ \midrule
\multirow{3}{*}{ Human}  
 &SLOPER4D                                                                     &      3.82                 &  10.39                   &        7.11              & 20.62          \\ 
& CityWalkers($50\%$)                                                              &    1.13                  &  3.34                          &      1.65                &      6.18   \\
& CityWalkers($100\%$)                                                      &      \textbf{1.09}                & \textbf{3.24}                       &  \textbf{1.61}                    &   \textbf{5.95}                    \\  

\bottomrule
\end{tabular}

\end{subtable}
\, \,  \,
     \begin{subtable}[t]{0.48\textwidth}
 
       \caption{Ablation on model components of PedGen.}
        \label{tab:ab_b}

 \begin{tabular}{@{}cccccc@{}}
\toprule
 \multirow{2}{*}{\begin{tabular}[c]{@{}c@{}}\textbf{Context}\\ \textbf{Factor}\end{tabular}} & \multirow{2}{*}{\textbf{Method}} & \multicolumn{4}{c}{\textbf{Metric} }                     \\  \cmidrule(l){3-6}  
&          & mADE $\downarrow$       & aADE $\downarrow$ & mFDE $\downarrow$ &  aFDE $\downarrow$      \\ \midrule
  \multirow{3}{*}{No} & PedGen (-traj/geo)                    &  1.33            &    4.47             &   1.97     &  8.29  \\
 & PedGen (-sep. token)             & 1.32            &       4.26        &  2.01      &   7.95  \\

 &      PedGen               &      \textbf{1.13}        &  \textbf{4.08}              &       \textbf{1.61}     &     \textbf{7.56}       \\ \midrule
      
       \multirow{3}{*}{Scene} &     PedGen (-3D rep.) &        1.13        &      4.26      & 1.60                   &   7.95 \\
      
           &   PedGen  (-semantic)                 &  1.12           & 3.86              &         1.56    &  7.09            \\
       &   PedGen   &   \textbf{1.11} & \textbf{3.75} &  \textbf{1.55} & \textbf{6.92}         \\

   \bottomrule
\end{tabular}

      \end{subtable}
      
\end{table}

\section{Conclusion}
 
\label{sec:conclusion}

We study a new task of generating context-aware pedestrian movements by learning from web videos with noisy labels. To facilitate this study, we collect a large-scale dataset CityWalkers with diverse real-world pedestrian movements in urban scenes. We further propose PedGen, a generative model that addresses the noisy labels in CityWalkers and models the three important context factors: the scene context, the personal characteristics, and the goal position. Experiments show that PedGen can generate realistic context-aware pedestrian movements in both real-world and simulation environments. We hope this study will present new opportunities and facilitate future research on modeling pedestrian movements in real-world settings.

\textbf{Limitations.} PedGen only considers static scene context at the starting frame, while pedestrian movement also depends on dynamic scene contexts such as the history trajectories of other pedestrians. Modeling these dynamic objects would be an interesting future direction. In addition,  PedGen only generates the movements of a single pedestrian at one time, while modeling group activities in urban scenes can help generate more realistic real-world behaviors.

\newpage

\bibliography{iclr2025_conference}
\bibliographystyle{iclr2025_conference}

\newpage

\centerline{\Large{\textbf{Appendix}}}
\appendix
We present additional details and experiments 
of the proposed CityWalkers dataset and the PedGen model for pedestrian movement generation. In Sec.~\ref{supp_data}, we introduce more details and statistics of CityWalkers.  Sec.~\ref{supp:benchmark} benchmarks the noise level of our automatic labeling pipeline. In Sec.~\ref{supp: model}, we discuss implementation details of PedGen. Sec.~\ref{sec:vis_citywalkers} provides more visualizations of the dataset and the qualitative results. We discuss important licensing and privacy considerations in Sec.~\ref{sec:license} and broader impacts of our work in Sec.~\ref{sec:impact}. More qualitative results can be found in the supplementary video. 
 
\section{More Details on CityWalkers}
\label{supp_data}

We manually review all videos to ensure the raw data quality and scene diversity. 
We collected 728 such videos and split each one into 5-second clips, with an interval of 25 seconds between subsequent clips.
We then perform a second check round to examine whether all clips are in urban environments, have proper lighting conditions without much motion blur, and do not contain abrupt viewpoint changes.
To prevent the leaking of personally identifiable information, we also blur the faces and license plates with mosaicing tools~\citep{xu2020centerface}.
In total, 22,698 clips are collected, each representing a different urban scene.
All video clips are decomposed into image sequences with a frame rate of 30 fps.
We keep pedestrians tracked consecutively for at least 10 frames, and pass it together with the camera angular velocity predicted by  DPVO~\citep{teed2023deep} to the WHAM network to get pseudo-labels for 4D global pedestrian movement.
For each pedestrian, WHAM outputs both its movement in the global frame $\mathcal{X}^g=\{\boldsymbol{t}_{t}^g, \boldsymbol{\phi}_{t}^g, \boldsymbol{\theta}_{t}, \boldsymbol{\beta}_{t}\}_{t=1}^{T}$, and in the local camera frame $\mathcal{X}^c=\{\boldsymbol{t}_{t}^{c(t)}, \boldsymbol{\phi}_{t}^{c(t)}, \boldsymbol{\theta}_{t}, \boldsymbol{\beta}_{t}\}_{t=1}^{T} $, note that the local camera frame $c(t)$ is varying with time as the camera is also moving in the video. To align the context information in the local camera frame at a specific timestep $\tau$ and the future pedestrian movement label in the global frame, we define an additional transformation matrix $\mathbf{T}_g^{c(\tau)} = [\mathbf{R}_g^{c(\tau)} | \mathbf{t}_g^{c(\tau)}]$ with $\mathbf{R}_g^{c(\tau)} = \boldsymbol{\phi}_\tau^{c(\tau)} (\boldsymbol{\phi}_\tau^{g})^{-1}$, $\mathbf{t}_g^{c(\tau)} = \boldsymbol{t}_\tau^{c(\tau)}  - \boldsymbol{\phi}_\tau^{c(\tau)} (\boldsymbol{\phi}_\tau^{g})^{-1}  \boldsymbol{t}_\tau^{g}$ and apply $\mathbf{T}_g^{c(\tau)}$ on the global human motion $\mathcal{X}^{g}$. The result is the global pedestrian movement in the fixed local camera coordinate $c(\tau)$, which can be paired with the context information as $\{\mathcal{I}_\tau, \mathcal{I}^d_\tau, \mathcal{I}^s_\tau,\mathcal{X}^{c(\tau)} \}$ to get training and validation samples. We use the CityScapes~\citep{cordts2016cityscapes} classes for the semantic map, as they contain common classes in urban scenes, such as buildings, sidewalks, and cars.

\begin{figure}[t!]
    \centering
    \includegraphics[width=1\linewidth]{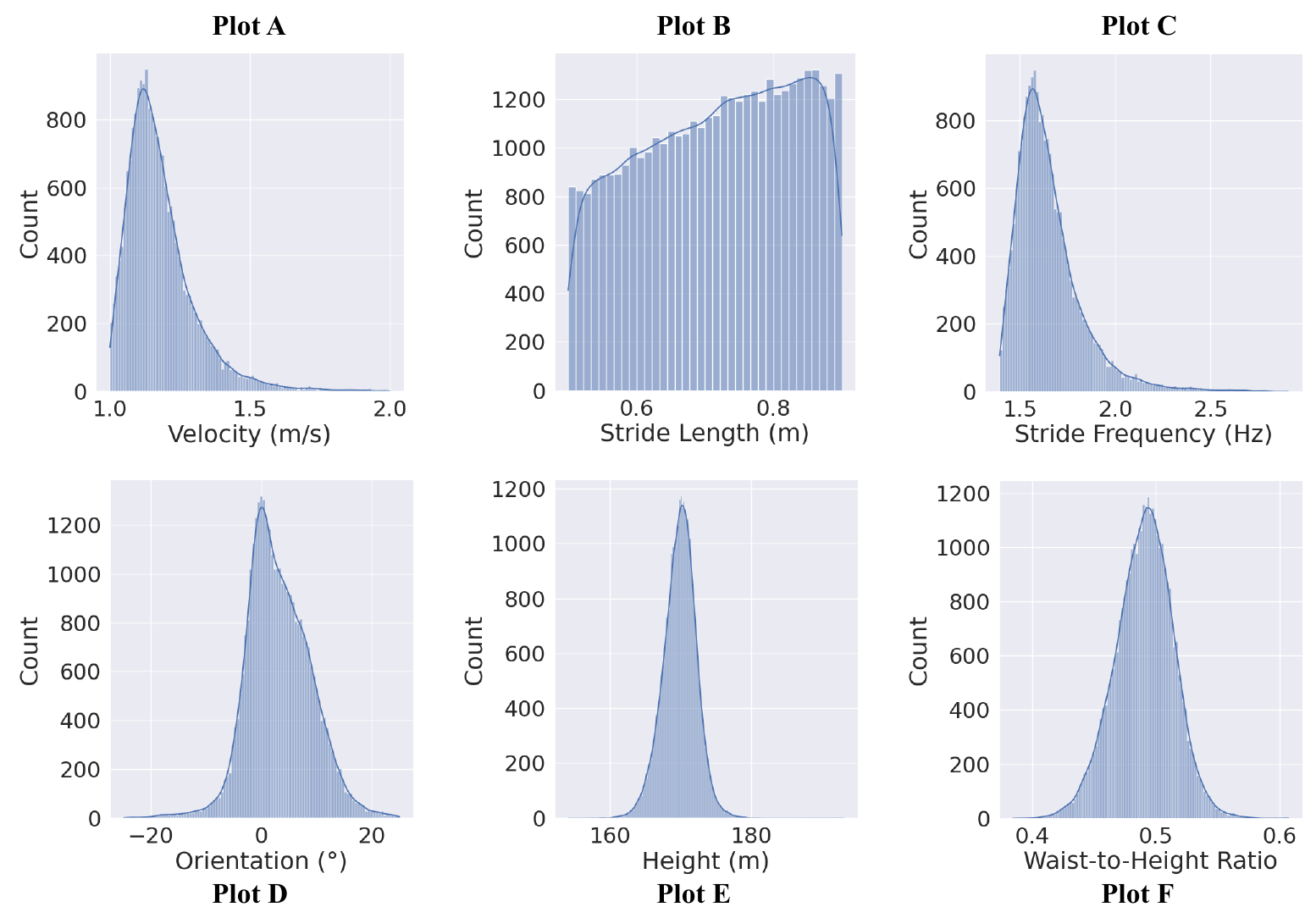}
    \caption{\textbf{Pedestrian movement statistics of CityWalkers.}}
    \label{fig:distribution}
\end{figure}

\begin{figure}[htbp]
    \centering
    \includegraphics[width=1\linewidth]{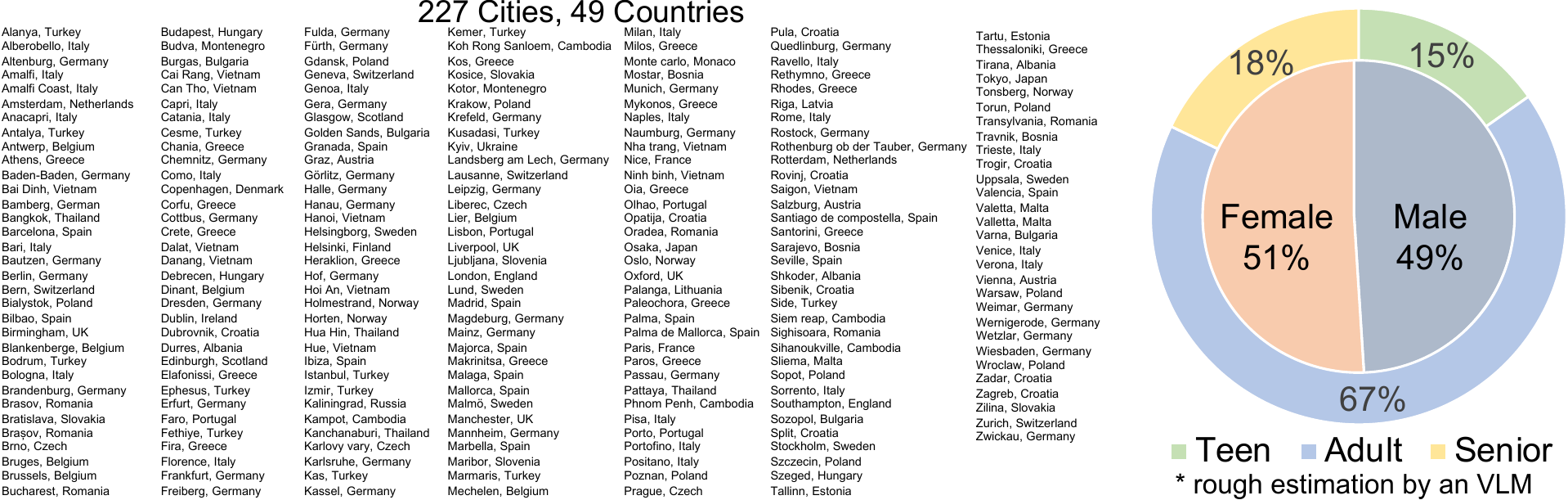}
    \caption{\textbf{List of locations and pedestrian attribute statistics of CityWalkers.}
     }
    \label{fig:distribution_data}
\end{figure}

Table~\ref{tab:dataset_compare} compares CityWalkers to other human motion datasets. CityWalkers has the most diverse human subjects and scenes compared to other human motion datasets and is the only dataset that uses web source videos and pseudo-labels. We provide further statistics regarding the pedestrian movements in CityWalkers in Fig.~\ref{fig:distribution}. Plots A-D display key motion characteristics. Plot A shows that our dataset captures motions with a wide range of typical human walking speeds. As evidenced by Plot B and C, our data also contains substantial samples of varying stride patterns. We also demonstrate the diversity of movement directions with Plot D, which represents the change in orientation across the recorded motion. In addition, we plot pedestrian body shape statistics with Plot E and Plot F. We look at the height of pedestrians in Plot E and their waist-to-height ratio in Plot F, as an indicator for the mass index.  Fig.~\ref{fig:distribution_data} demonstrates the list of cities and countries in CityWalkers and its pedestrian attributes roughly estimated by an off-the-shelf VLM~\citep{chen2023internvl}. CityWalkers covers most European countries and some Asian countries, and we plan to add more locations in the future. As most of the places in CityWalkers have many tourists, its pedestrians are from all over the world, and the age groups and genders are well-represented.

\begin{table}[t!]
\centering

\setlength\tabcolsep{2.0pt}
\scriptsize

\caption{\textbf{Comparison of human motion datasets}. We compare CityWalkers with datasets that do not have scene context (top), datasets that provide scene context labels but are captured in controlled environments (middle), and datasets that are captured in the wild (bottom). CityWalkers has the most number of scenes and subjects among all datasets and is the only dataset that uses web source videos and pseudo-labels.
}
\begin{tabular}{@{}lcccccccccc@{}}
\toprule
Datasets &Web source  &  Pseudo-labels &  \# Subjects & \# Scenes & Hours & Depth                                         & Segmentation              & SMPL   \\ \midrule
   AMASS~\citep{mahmood2019amass}      & -&  -& 344         &       -    &          62.9        &     -                                &                  -          &    \ding{51}   
   \\

     DNA-Rendering~\citep{cheng2023dna}      &  - &  -&  1,500       &      -    & 3.2    &     -                                               &              -         &     \ding{51}  \\ \midrule
                 


                                 PROX~\citep{hassan2019resolving}       & - &   -&   20       &      12     & 0.9     &     \ding{51}                                                 &                 \ding{51}           &     \ding{51}                 \\ 
                                     RELI11D~\citep{yan2024reli11d}     & -  &  -&   10       &   7        & 3.3     &        \ding{51}                                              &                  -          &       \ding{51}            \\ 
                                  RICH~\citep{huang2022capturing}     & -  &  -&   22       &   5        & 2.7      &        \ding{51}                                              &                  -          &       \ding{51}            \\ 
                                    TRUMANS~\citep{jiang2024scaling}      &  - & -&   7       &      100     & 15.0     &     \ding{51}                                                 &                 \ding{51}           &    \ding{51}   \\ \midrule

    3DPW~\citep{von2018recovering}  & - &  -&7 & 60 & 0.5 &-&-&\ding{51} \\

      EMDB~\citep{kaufmann2023emdb}  & - &  -&10 & 81 & 1.0 &-&-&\ding{51} \\
 SLOPER4D~\citep{Dai_2023_CVPR}  & - &  -&12 & 10 & 1.4 &\ding{51}&-&\ding{51} \\

               JRDB-Pose~\citep{vendrow2023jrdb}  & - & -& 5,022 & 54 & 1.1 &\ding{51}&\ding{51}&-  \\




   \midrule
   
\textbf{CityWalkers (ours)}   & \ding{51} & \ding{51} & \textbf{120,914}       & \textbf{16,215}     & 30.8  & \ding{51}  & \ding{51} &\ding{51}              \\ \bottomrule
\end{tabular}
 
\label{tab:dataset_compare}
\end{table}

\section{Benchmarking the Noise Level of CityWakers}
\label{supp:benchmark}
Though we have applied several techniques to improve the quality of pseudo-labels by using state-of-the-art models and filtering wrong predictions, label noise from web videos is still inevitable and hence it is important to benchmark the accuracy of our data autolabeling pipeline.  We use the SLOPER4D dataset~\citep{Dai_2023_CVPR}, collected in a similar outdoor setting as CityWalkers with a much smaller scale. The SLOPER4D dataset has ground-truth human motion and scene-depth labels annotated from the 3D LiDAR point clouds. We evaluate the accuracy of the 4D human motion estimation with the Procrustes-Aligned Mean Per Joint Position Error (PA-MPJPE) and the World-Aligned Mean Per Joint Position Error (WA-MPJPE), and the accuracy of the monocular depth estimation with the absolute relative error (REL) and the Root Mean Squared Error (RMSE). The results are shown in Tab.~\ref{tab:benchmark}. We want to stress that while encountering label noise is inevitable, web videos are necessary to learn a generalizable pedestrian movement generation model with natural pedestrian behaviors and diverse motion contexts. 

\begin{table}[h]
\centering

\scriptsize
\caption{\textbf{Benchmarking results of our autolabeling pipeline on the SLOPER4D dataset.}} 
\begin{tabular}{@{}cc|cc@{}}
\toprule
 \multicolumn{2}{c}{4D Human Motion} & \multicolumn{2}{c}{Depth} \\ \midrule
PA-MPJPE (mm)$\downarrow$   & WA-MPJPE (mm) $\downarrow$  & REL    $\downarrow$     & RMSE  $\downarrow$      \\
 42.76            & 297.72           & 0.33        & 5.46     \\   \bottomrule
\end{tabular}
\label{tab:benchmark}
\end{table}

\section{Implementation Details of PedGen}
\label{supp: model}
The denoising transformer of PedGen follows the decoder-only architecture with 8 transformer blocks and residual connections between them, and the latent dimension of the transformer is 512. Each transformer block consists of a self-attention and an MLP module with a FiLM~\citep{perez2018film} layer applied after each module to inject the condition information similar to~\citep{chen2023humanmac}.  During training, we found our depth labels could be inconsistent with the depth estimated from the SMPL global translation in the dataset. Therefore, we multiply the depth map by a factor $\gamma$, which equals the ratio between the depth from the SMPL root translation and the depth of the human root's projection in the 2D depth label to align the motion and the scene. To ensure the generated motion starts from the initial position $\boldsymbol{t}_1$ and ends at the goal position $\boldsymbol{t}_T$ when the goal condition is provided, we apply diffusion inpainting~\citep{chen2023humanmac} by setting $\boldsymbol{\hat{v}}_1=\boldsymbol{0
}$ and scale the velocity predictions  $\boldsymbol{\hat{v}}_t = \lambda \boldsymbol{\hat{v}}_t$ by $\lambda= (\boldsymbol{t}_T-\boldsymbol{t}_1)/\hat{\boldsymbol{t}_T}$. Our model also supports the efficient generation of long-term movements by concatenating the short-term movements generated by PedGen. Specifically, we define a transition phase between two adjacent motion intervals. During the initial generation, we force the motion of the transition phase to be the same in adjacent intervals and then partially add noise to the transition phase again and denoise it back to generate a smoother transition similar to~~\citep{shafir2023human}. Please see the attached video for results on long-term movement generation.

All models are trained with 4 Nvidia RTX A5000 GPUs with 256 batch size. We use the Adam~\citep{kingma2014adam} optimizer with a weight decay of 1e-7 and a gradient clipping of 1.0 to train for 500 epochs on CityWalkers. The initial learning rate is set to 4e-4, and it decays by a factor of 0.9 every 75 epochs. The whole model took about 30 hours to train. The number of forward diffusion steps is set to $K=1000$  with a cosine noise schedule. The loss weights $w_{\mathrm{rec}}, w_{\mathrm{traj}}, w_{\mathrm{geo}}$ are all set to 1.   We apply data augmentation during training, which randomly rotates the pedestrian movement together with its scene context encoded in the local voxel along the vertical axis.  We further use classifier-free guidance~\citep{ho2022classifier} with a $20\%$ probability of dropping the condition embedding during training, and the guidance scale is set to $1.0$ during inference. We also adopt the DDIM sampler~\citep{song2020denoising} with 100 steps to speed up sampling.  As the original videos do not have ground truth camera intrinsic parameters, we estimate them by setting the focal length to be the diagonal pixel length of the image and the optical center to be the center of the image. For the local point cloud $\mathcal{P}_{\mathrm{local}}$, we set its range $\Delta x= 4\mathrm{m}$, $\Delta y = 2\mathrm{m}$ and $\Delta z = 4\mathrm{m}$, the grid size for voxelization is further set to be $0.2\mathrm{m}\times0.1\mathrm{m}\times0.2\mathrm{m}$, resulting in a voxel shape of $40\times40\times40$.  To encode the voxel, we first treat the vertical axis (Y-axis) as the feature dimension and the rest of the axes as the tokens. We then use a cross-attention layer to extract information from the encoded scene with a query vector from the other context factors. 

\section{Additional Visualizations}
\label{sec:vis_citywalkers}
\begin{figure}[htbp]
    \centering
    \includegraphics[width=1.0\linewidth]{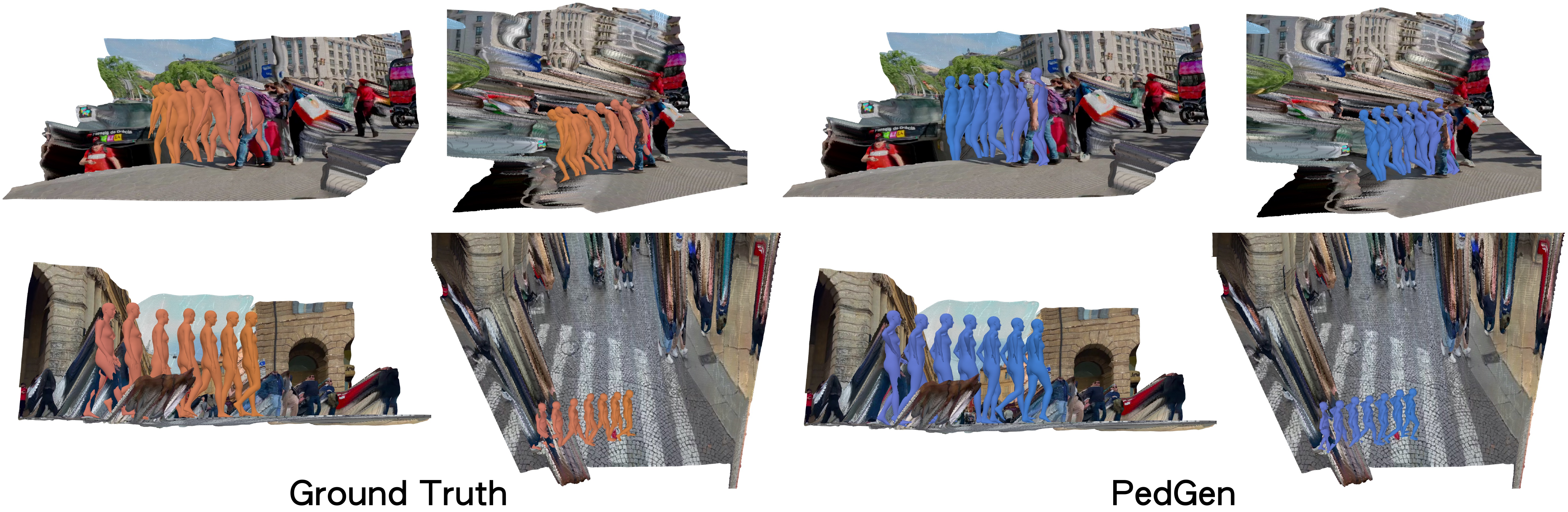}
    \caption{\textbf{Visualizations in 3D.} We visualize both the ground truth scene and movement labels (orange) and the generated movements by PedGen (blue) in multiple views in 3D by unprojecting the image pixels from the depth labels.
    }
    \label{fig:vis_3d}
\end{figure}

\begin{figure}[htbp]
    \centering
    \includegraphics[width=1.0\linewidth]{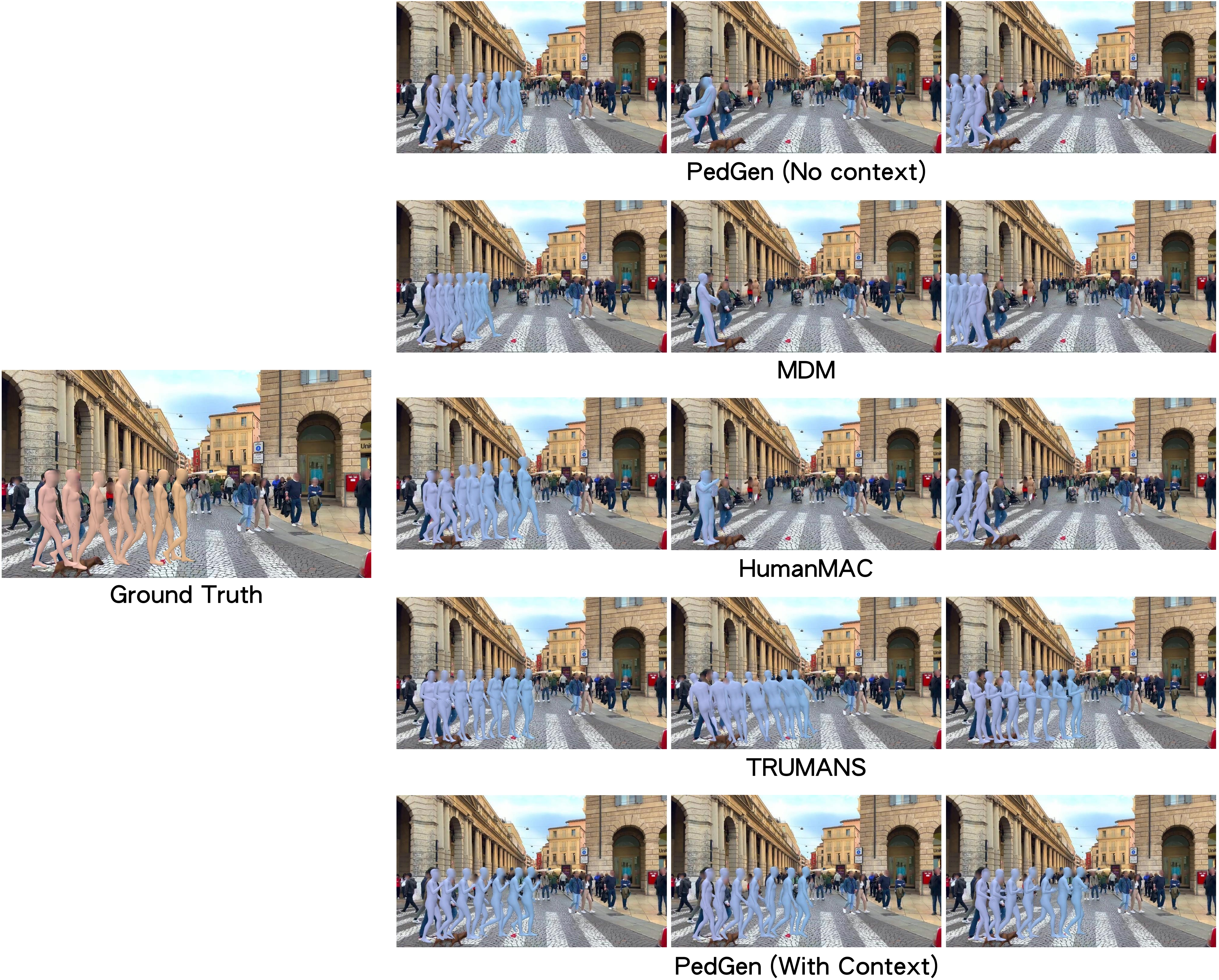}
    \caption{\textbf{Qualitative comparison results.} We visualize the generation results of PedGen compared to the other baselines and the ground truth.  Three random samples are generated for each method.
    }
    \label{fig:vis_compare}
\end{figure}

\begin{figure}[H]
    \centering
    \includegraphics[width=1\linewidth]{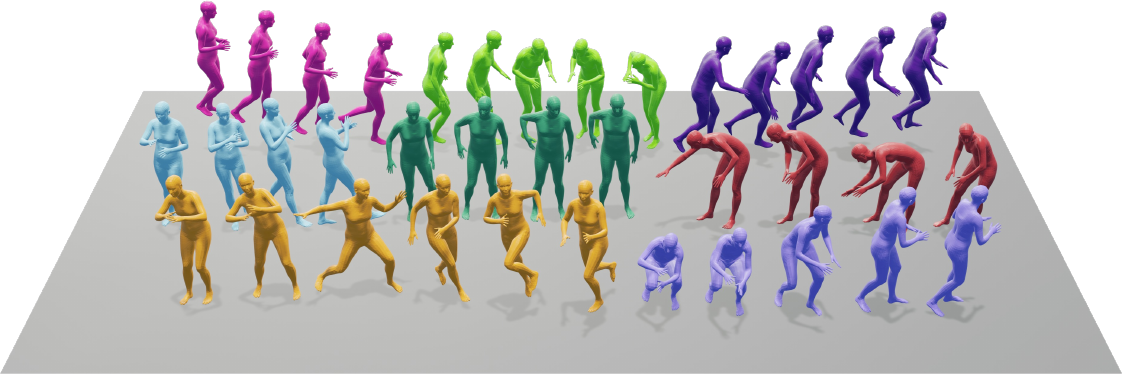}
    \caption{\textbf{Samples of 4D pedestrian movement labels in CityWalkers.} The text descriptions of the movements from top left to bottom right are: walking down stairs (pink), turning and lifting baggage up steps (light green), walking up stairs (dark purple), turning around with phone in hand (sky blue), moving hands to hip (dark green), wiping seats and tables (red), jumping and skipping around (yellow), taking photo and standing up (light purple). }
    \label{fig:additional_movement}
\end{figure}

\begin{figure}[htbp]
    \centering
    \includegraphics[width=1\linewidth]{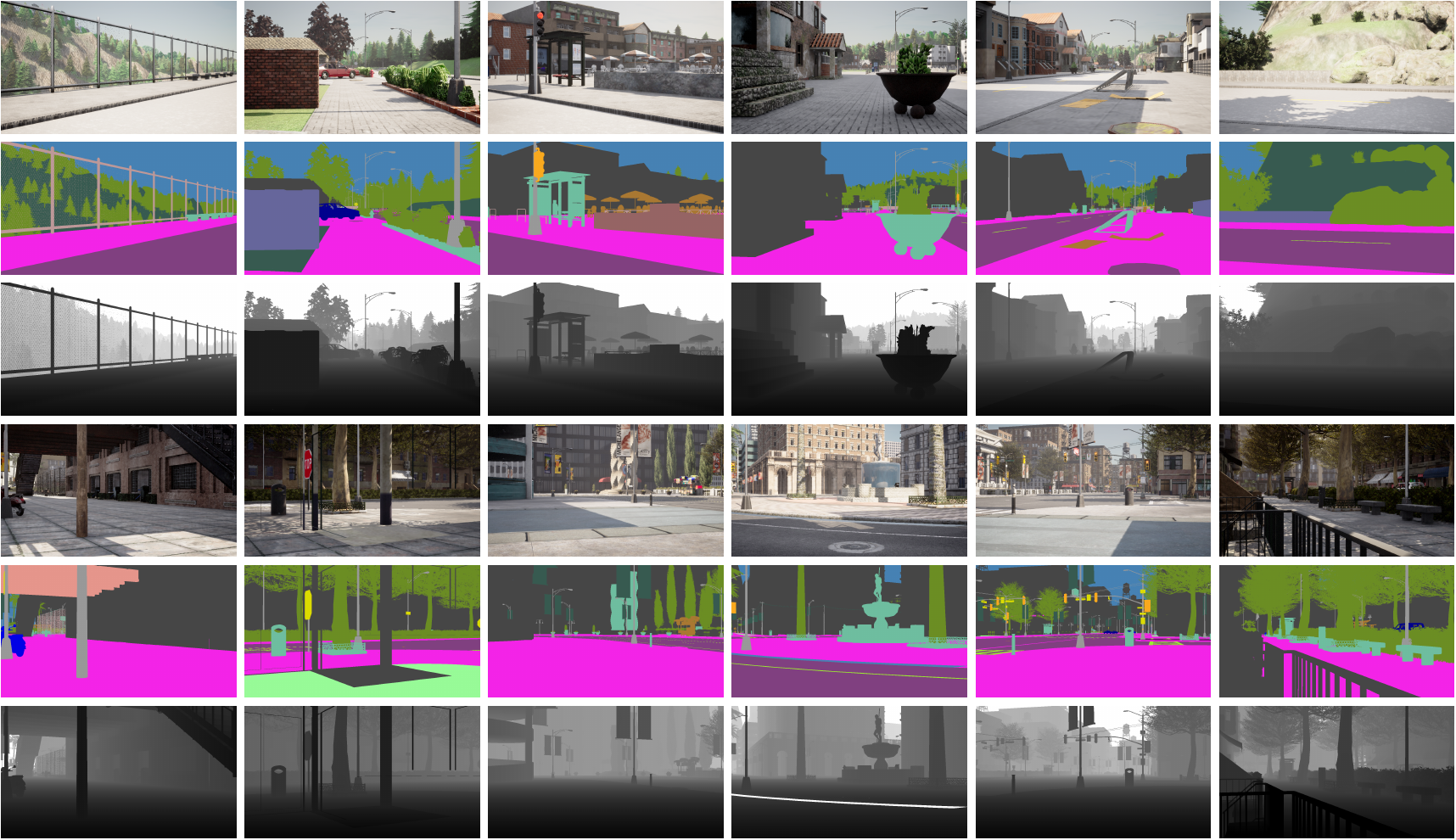}
    \caption{\textbf{Samples in the CARLA test set.} Each scene contains a rendered image, a semantic map, and a depth map.}
    \label{fig:vis_diverse_carla}
\end{figure}

\begin{figure}[h!]
    \centering
    \includegraphics[width=1\linewidth]{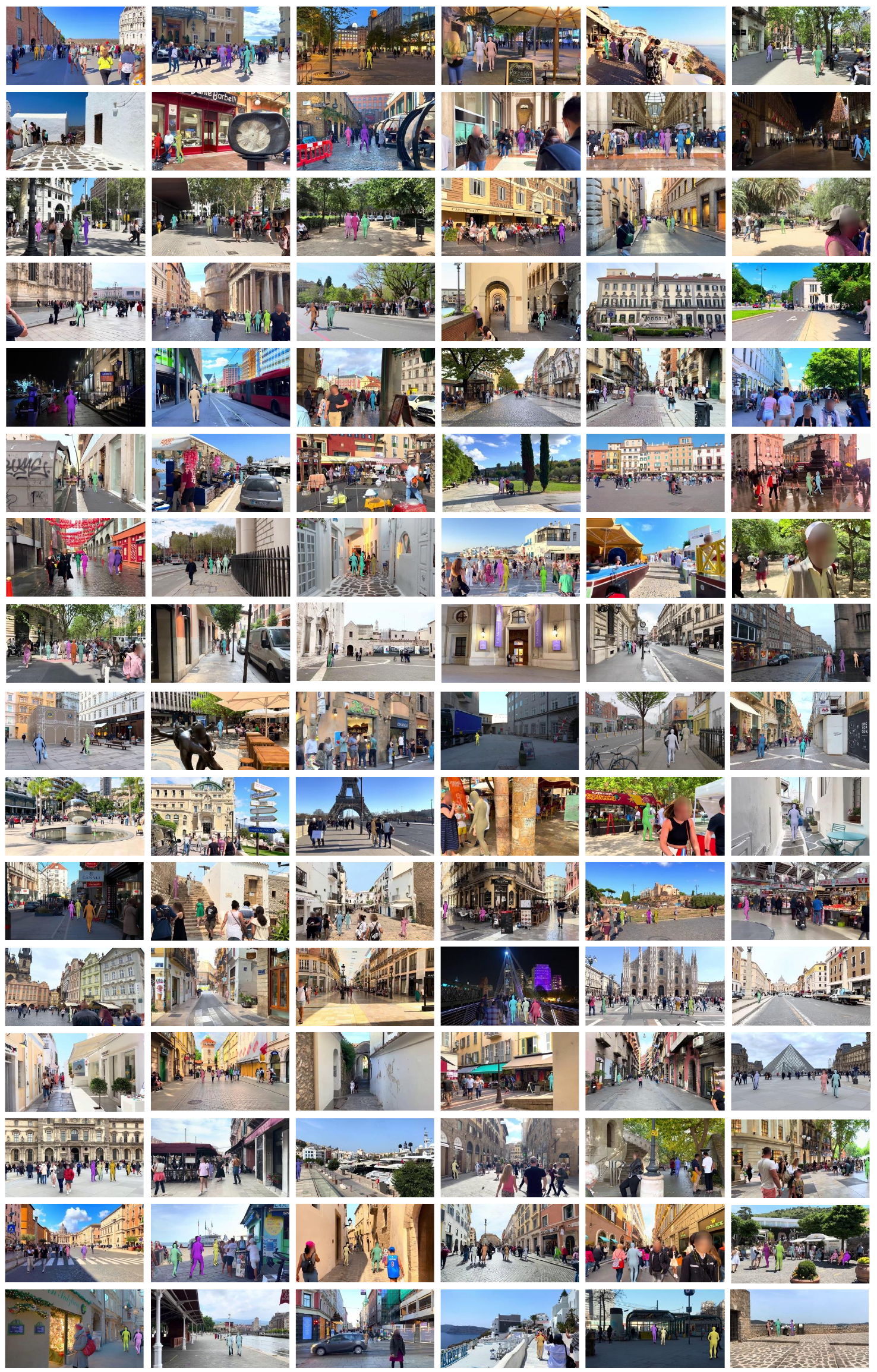}
    \caption{\textbf{Samples of real-world scenes in CityWalkers.} We visualize the extracted pedestrian 3D meshes in each scene.}
    \label{fig:vis_scene}
\end{figure}

\begin{figure}[htbp]
    \centering
    \includegraphics[width=0.9\linewidth]{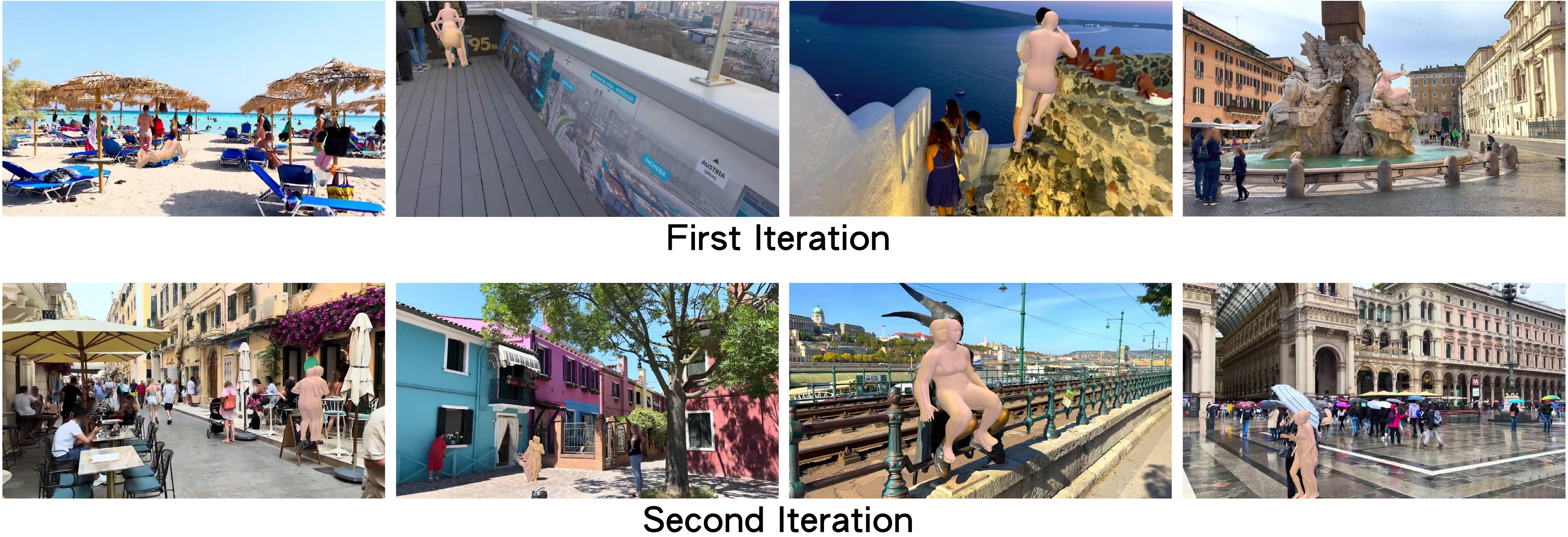}
    \caption{\textbf{Visualizations of anomaly labels in CityWalkers.} We visualize the filtered labels in the first and the second iterations of automatic anomaly label filtering.
    }
    \label{fig:vis_anomaly}
\end{figure}

Figure~\ref{fig:vis_3d} shows the 3D visualizations of the ground truth and generated pedestrian movements of PedGen, as well as the scene context labels by unprojecting the image pixels to the 3D space according to the depth label. We can see that the ground truth movements align well with the geometry of the scene, showing the quality of the movement and scene context labels. The generated pedestrian movements also match well with their surrounding environments in 3D. Fig.~\ref{fig:vis_compare} shows a qualitative comparison between PedGen and other baselines. It can be noticed that PedGen can generate more natural and diverse motions with different poses, gestures, and hand movements. It also aligns with the surrounding environment better after being conditioned on the context factors.   Some additional 4D pedestrian movement labels in CityWalkers are visualized in Fig.~\ref{fig:additional_movement} to show the diversity of the pedestrian movements.  Fig.~\ref{fig:vis_diverse_carla} shows samples of the CARLA test set that we used to evaluate the zero-shot generalization ability of PedGen. We display a handful of urban scenes with diverse objects and layouts, all of which reflect test set diversity.  In Fig.~\ref{fig:vis_scene}, we lay out a wide array of scenes in CityWalkers with diverse locations, weather, crowd density, and time of day to show the diversity of the urban scenes. Fig.~\ref{fig:vis_anomaly} shows visualizations of the automatically filtered anomaly samples. We can observe that the anomaly labels in the first iteration of anomaly filtering have drastic errors in the body pose or do not belong to pedestrian movements. In contrast, the anomaly labels in the second iteration of anomaly filtering have much smaller errors with minor deviations in the local movements and could also contain false positives with novel movements. We find that two iterations with a reconstruction error threshold of 10 are sufficient to filter most low-quality labels with the best model performance. An ablation study on the number of filtering iterations can be found in Tab.~\ref{ab:filtering}.

\section{Ethics and Privacy Considerations}
\label{sec:license}

We will follow practices in the existing YouTube datasets~\citep{abu2016youtube, xu2018youtube} for privacy protection on web videos. All of our raw videos~\citep{youtube_poptravel} have a Creative Commons license (CC-BY-SA\footnote{\url{https://creativecommons.org/licenses/by/3.0/legalcode.en}}) and are complied with YouTube’s privacy policy and terms of service~\footnote{\url{https://www.youtube.com/static?template=terms}}.
Besides, we skip copyright-related information during data processing to protect the rights of logos, channel owner information, or other copyrighted materials. We will not provide processed video clips, and users are redirected to original YouTube videos via a link and follow our provided pre-processing scripts to process the dataset themselves.

Our data will be released under the CC BY-NC-SA 4.0 license\footnote{\url{https://creativecommons.org/licenses/by-nc-sa/4.0/deed.en}}.  To protect privacy, we remove all personally identifiable information by adding mosaics to faces and license plates in the videos following~\citep{grauman2022ego4d}. We will also implement safeguards on our dataset webpage with detailed user agreements and rules to encrypt personal information, enforce access limits, and monitor misuse and unauthorized data access. Upon data release, we will credit the source, provide a link to the license, and state that no modifications have been made to the raw videos except for the appended pedestrian movement and scene labels, and that the data shall not be used for commercial purposes. We will follow the guidelines of our institute's institutional review board and comply with applicable laws. For instance, the human subjects in the videos have the right to view, correct, and delete personal information in the dataset. The review board will also evaluate the collection, use, and sharing of the data to ensure alignment with best practices in data privacy and ethics.

\section{Broader Impacts}
Our work can benefit many applications. For example, city designers can simulate pedestrian movements to optimize public areas and transportation systems. Forecasting future pedestrian movements is also crucial for the safe deployment of autonomous vehicles. The diverse urban scenes and pedestrians in the CityWalkers dataset could also support future research directions besides pedestrian movement generation, such as relation modeling between real-world scenes and humans and integration of realistic human movements and urban scenes into embodied AI training.  

Some potential negative societal impacts of our work include using the released data for surveillance applications by modeling and predicting pedestrian behaviors and the risk of leaking personally identifiable information. It is also possible that our model could be misused to generate fake pedestrian movements of real-world humans. As discussed in Sec.~\ref{sec:license}, we will enforce practices to protect privacy (e.g., removing personally identifiable information) and add the user agreement and the license to our dataset to prevent misuse.

\label{sec:impact}


\clearpage

\begin{table}[htbp] 
\scriptsize
\centering
\caption{\textbf{Ablation on the number of filtering iterations.} We evaluate PedGen with no context on the CityWalkers validation set.}
     \label{ab:filtering}
\begin{tabular}{@{}ccccc@{}}
\toprule

\multirow{2}{*}{\begin{tabular}[c]{@{}c@{}}\textbf{Filtering} \\ \textbf{Iterations}\end{tabular}}  & \multicolumn{4}{c}{\textbf{Metric}}   \\ \cmidrule(l){2-5}                                                                                & mADE $\downarrow$   & aADE $\downarrow$   & mFDE $\downarrow$  & aFDE $\downarrow$    \\ \midrule
0                                                                                                                                         &     1.17   &  4.45      &    1.64   & 8.31         \\
1                                                                        &    1.17     &  \textbf{4.22}      &  1.68      &    \textbf{7.88}      \\
2                                                                          & \textbf{1.13}       &    4.32     & \textbf{1.60}       &    8.09      \\
3                                                                  & 1.17  &  4.42       &    1.63    &  8.22                 \\
  \bottomrule
\end{tabular}
\end{table}

\end{document}